\documentclass[runningheads]{llncs}

\usepackage{accv}

\usepackage{accvabbrv}

\usepackage{graphicx}
\usepackage{booktabs}
\usepackage{float}
\usepackage{colortbl}
\usepackage{graphicx}
\usepackage{subcaption}
\usepackage{multirow}
\usepackage{wrapfig}
\usepackage{enumitem}
\usepackage{amsmath}
\usepackage{amssymb}
\usepackage{mathtools}

\usepackage[accsupp]{axessibility}  

\usepackage[pagebackref,breaklinks,colorlinks,citecolor=accvblue]{hyperref}

\usepackage{orcidlink}

\begin{document}
 
\title{Simple Supervision Is Hard to Beat: \\A Bitter Lesson from Sparse Target Labels in Domain-Adaptive Object Detection} 

\titlerunning{Random Target Labels for SFDA-OD}

\author{Lijun Zhang\inst{1}\orcidlink{0000-0003-1946-7105} \and
Ruinian Xu\inst{1} \and
Mudit Agrawal\inst{1}}

\authorrunning{Lijun Zhang, Ruinian Xu, et al.}

\institute{
Amazon Robotics, Seattle, USA\\
\email{\{ljzhang,ruinianx,muditagr\}@amazon.com}}

\maketitle

\begin{abstract}
  Source-free domain adaptive object detection adapts a source-trained detector to an unlabeled target domain, typically through teacher-student self-training with pseudo-labels. 
  We revisit this setting when a small, uniformly sampled subset of target images is labeled. 
  We introduce Random-Target Supervised Mixing (RTSM), a simple anchor that incorporates these annotations through a supervised detection loss while leaving the original unlabeled adaptation branch unchanged. 
  Across evaluations spanning four SFDA-OD methods, two object detectors, multiple adaptation tasks, and target-label budgets from $1\%$ to $10\%$, RTSM consistently improves pure SFDA by $1.7$ to $18.3$ AP50. 
  We then examine whether the same annotations can provide further gains by steering unlabeled self-training. 
  To this end, we evaluate ten sparse-label feedback plugins covering pseudo-label selection, object completion, and optimization control, which yield limited and method-dependent gains over RTSM. 
  These results reveal a bitter lesson for sparse-label SFDA-OD: simple supervision is hard to beat. 
  RTSM therefore provides a simple yet effective anchor for sparse-label SFDA-OD.
\end{abstract}

%
%

\section{Introduction} \label{sec:intro}
%
%
%
Object detectors are often trained once and then deployed in environments whose visual statistics may differ from the training data.
In autonomous driving and surveillance scenarios, for example, shifts in weather, illumination, camera hardware, and geographic layout can substantially degrade a detector trained on a labeled source domain \cite{Chen_2018_CVPR,Saito_2019_CVPR}.
Source-free domain adaptive object detection (SFDA-OD) addresses this problem by adapting a source-trained detector to an unlabeled target domain without accessing the source data \cite{Li_2021_AAAI_FreeLunch}.

%
%
%
Recent SFDA-OD methods adapt the detector through self-training.
A teacher model predicts pseudo labels for unlabeled target images, and a student model is optimized to match those pseudo labels, often within a Mean Teacher framework that stabilizes training through exponential moving average updates \cite{Tarvainen_2017_NeurIPS}.
This paradigm underlies many strong SFDA-OD methods, such as SED \cite{Li_2021_AAAI_FreeLunch}, LPU \cite{Chen_2023_ACMMM_LPU}, PETS \cite{Liu_2023_ICCV_PETS}, LPLD \cite{Yoon_2024_ECCV_LPLD}, and DDT \cite{He_2025_ICCV_DDT}.

However, the pseudo-label dependence also creates a bottleneck.
Under target-domain shift, pseudo labels can include false positives, miss hard target objects, and provide unstable optimization signals.
Existing methods address parts of this problem by refining confidence thresholds, exploiting low-confidence proposals, stabilizing teacher updates, or exchanging teacher and student models \cite{Chen_2023_ACMMM_LPU,Liu_2023_ICCV_PETS,Yoon_2024_ECCV_LPLD,He_2025_ICCV_DDT}.
Active domain adaptation studies a related question by selecting a small subset of target samples to annotate as more reliable signals for domain adaptation \cite{Nakamura_2024_CVPR_ADA_FN,Menke_2024_ESWA_ADAOD}.
These strategies are valuable to select the most important samples for annotation, but their performance can depend on which proxy is chosen and how well it transfers under domain shift.

%
We therefore study a simpler protocol where the labeled target subset is selected uniformly at random.
Random selection is not meant to be the most label-efficient annotation strategy.
Instead, it removes the active-selection proxy from the problem and lets us isolate a different question.
Once a small set of human target labels is given, how should those labels be used to improve existing SFDA-OD methods?

%
%
%

\begin{wrapfigure}{r}{0.42\linewidth}
    \vspace{-25pt}
    \centering
    \includegraphics[width=\linewidth]{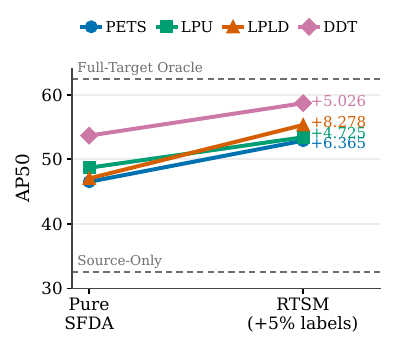}
    \caption{RTSM gives a strong supervised-mixing anchor across the evaluated SFDA-OD methods.
    Dashed lines mark the source-only and full-target oracle references.}
    \label{fig:intro_rtsm_anchor}
    \vspace{-20pt}
\end{wrapfigure}

The most direct answer is to add a supervised detection loss on the randomly labeled target images while keeping the original SFDA-OD training procedure unchanged.
We call this baseline \emph{Random-Target Supervised Mixing} (RTSM), which treats sparse target labels as ordinary supervised examples and mixes them with the unlabeled self-training branch of an existing SFDA-OD method.
RTSM is algorithmically simple, but it is not a weak control.
As shown in Figure~\ref{fig:intro_rtsm_anchor}, RTSM substantially improves the source-free baseline detectors.
For instance, for DDT~\cite{He_2025_ICCV_DDT}, adding 5\% randomly labeled target data improves AP50 (average precision at IoU 0.5) from 53.669 to 58.695.
Therefore, rather than asking whether sparse labels help at all, we ask whether they can help \emph{beyond} RTSM by improving the unlabeled self-training branch.

%
%
%
To systematically study this research question, we consider three ways sparse target labels may interact with pseudo supervision.
\emph{Pseudo-label selection} treats the labeled subset as a target-domain reliability probe and adjusts which pseudo labels should be trusted.
Its main target is the false-positive side of pseudo-label noise, including unreliable confidence estimates.
\emph{Object completion} instead focuses on what the teacher fails to predict.
It uses the labeled subset to characterize missed objects and guide the revival of candidate pseudo labels, targeting false negatives and under-activated detector queries.
\emph{Optimization control} does not directly change the pseudo labels.
It uses sparse labels to regulate how supervised and pseudo-label losses jointly update the detector, targeting the training dynamics between trusted human labels and noisy pseudo labels.

%
%
%
Our study reveals a bitter lesson for sparse-label SFDA-OD: simple supervision is remarkably hard to beat. 
Across various SFDA-OD settings, sparse target annotations consistently improve adaptation when used directly through supervised mixing. 
Yet the same annotations provide much less reliable guidance when used to calibrate pseudo-labels, recover missed objects, or regulate optimization, with gains that remain fragmented and method-dependent. 
This distinction is particularly important for the emerging sparse-label SFDA-OD setting, which lies at the intersection of semi-supervised domain adaptation and active domain adaptive object detection~\cite{Saito_2019_ICCV_MME,Nakamura_2024_CVPR_ADA_FN,Menke_2024_ESWA_ADAOD}. 
Once target annotations are available, improvement over pure SFDA alone is no longer sufficient evidence of effective label utilization. 
RTSM therefore serves as the minimal and necessary anchor, isolating the benefit obtained from direct target supervision. 

%
Our main findings are summarized as follows.
\begin{itemize}[noitemsep,nolistsep,topsep=0pt,leftmargin=*]
\item Direct supervision provides a remarkably strong use of sparse target annotations.
Across evaluations spanning four SFDA-OD methods, two object detectors, multiple adaptation tasks, and target-label budgets from $1\%$ to $10\%$, RTSM consistently improves pure SFDA by $1.7$ to $18.3$ AP50.

\item The same annotations are considerably less reliable when used to steer self-training.
Across ten plugins covering pseudo-label selection, object completion, and optimization control, no approach consistently outperforms RTSM.

\item The inconsistency is systematic rather than confined to a particular experimental setting.
Plugin gains vary across object categories and adaptation pipelines, while their marginal benefit generally decreases as the labeled budget grows.
\end{itemize}

%


\section{Preliminary} \label{sec:prelim}
We first introduce the object detection and source-free domain adaptive object detection (SFDA-OD) required by our study.
We then define the RTSM anchor used throughout the paper.
Detailed related work is provided in Appendix~\ref{app:related}.

\textbf{Transformer-based Object Detector.}
Object detection maps an image to a set of object instances, each specified by a semantic category and a bounding box.
Classical neural detectors implement this mapping with proposal-based or dense prediction pipelines, such as Faster R-CNN and RetinaNet \cite{Ren_2015_FasterRCNN,Lin_2017_RetinaNet}.
Transformer-based detectors further cast detection as end-to-end set prediction, where a fixed set of object queries represents candidate objects and interacts with global image features \cite{Carion_2020_DETR,Zhang_2022_DINO}.

Formally, given an image $x$, we denote the detector output as $M$ object-query predictions
\begin{equation}
    f_{\theta}(x)=\{(\mathbf{p}_j,\mathbf{b}_j,\mathbf{q}_j)\}_{j=1}^{M},
\end{equation}
Here the three terms denote the class probability vector, the predicted bounding box, and the corresponding query representation.
For a labeled image $(x,y)$ with annotations $y=\{(c_k,\mathbf{b}_k)\}_{k=1}^{K}$, DETR-style detectors match predictions to ground-truth objects with bipartite matching.
The supervised detection loss after matching is denoted as $\mathcal{L}_{\mathrm{det}}(f_{\theta}(x),y)$, which consists of a classification term for category prediction and localization terms for box regression, such as $\ell_1$ and generalized IoU losses.
In Section~\ref{sec:method}, $\mathbf{p}_j$, $\mathbf{b}_j$, $\mathbf{q}_j$, and $\mathcal{L}_{\mathrm{det}}$ serve as the common detector interface for defining pseudo-label selection, object completion, and optimization control.

\textbf{Mean Teacher Self-Training in SFDA-OD.}
%
%
SFDA-OD aims to obtain a target-adapted detector when the source data are no longer accessible during adaptation.
Let $f_{\theta_s}$ denote the detector trained on the source domain, and let $\mathcal{D}_t^u=\{x_i^t\}_{i=1}^{N_t}$ denote the unlabeled target training set.
Starting from $f_{\theta_s}$, the adaptation procedure optimizes a detector on $\mathcal{D}_t^u$ and returns the target detector $f_{\theta_t}$.

As a widely-used framework, Mean Teacher Self-Training \cite{Tarvainen_2017_NeurIPS,He_2025_ICCV_DDT} instantiates this adaptation by using a temporally averaged model to generate pseudo supervision for unlabeled target images.
It maintains a student detector $f_{\mathrm{stu}}$ and an EMA teacher detector $f_{\mathrm{tea}}$ respectively.
Both networks are initialized from the source detector $f_{\theta_s}$.
The student is updated by gradient descent, and the teacher averages the student over training steps to suppress instantaneous optimization noise.
After adaptation, the trained student becomes the target detector $f_{\theta_t}$.

For a target image $x_i^t$, the teacher first predicts candidate detections on a weakly augmented view,
\begin{equation}
\mathcal{R}_i
= f_{\mathrm{tea}}(a_w(x_i^t))
= \{(\hat{c}_{ij},\hat{\mathbf{b}}_{ij},s_{ij},\mathbf{z}_{ij})\}_{j=1}^{M_i}, 
\end{equation}
where $a_w$ denotes a weak augmentation that preserves the image content with mild transformations~\cite{Sohn_2020_NeurIPS_FixMatch,Liu_2021_ICLR_UnbiasedTeacher}, and $M_i$ denotes the number of candidate predictions within the image, and the prediction outputs are the predicted class, predicted box, confidence score, and query representation, respectively.

A pseudo-label rule maps these candidates to accepted pseudo-labels, i.e., $\hat{y}_i = \Gamma(\mathcal{R}_i)$, where $\Gamma$ denotes the pseudo-label generation rule, such as confidence thresholds, non-maximum suppression, and low-confidence proposal handling \cite{Sohn_2020_NeurIPS_FixMatch,Liu_2023_ICCV_PETS,He_2025_ICCV_DDT}.
The pseudo-labels then supervise the student on a strongly augmented target view $a_s(x_i^t)$, where $a_s$ applies stronger photometric, geometric, or masking perturbations~\cite{Sohn_2020_NeurIPS_FixMatch,Cubuk_2019_RandAugment,Liu_2021_ICLR_UnbiasedTeacher}.
The full unlabeled branch is denoted by
\begin{equation}
\mathcal{L}_{\mathrm{u}}
=
\frac{1}{|\mathcal{B}_{u}|}
\sum_{x_i^t \in \mathcal{B}_{u}}
\mathcal{L}_{\mathrm{det}}\left(f_{\mathrm{stu}}(a_s(x_i^t)),\hat{y}_i\right)
+
\mathcal{L}_{\mathrm{aux}}.
\end{equation}
where $\mathcal{B}_u$ is an unlabeled target mini-batch.
Here the auxiliary term $\mathcal{L}_{\mathrm{aux}}$ represents any additional unlabeled loss used by the SFDA-OD baseline.
Different SFDA-OD methods instantiate the pseudo-label rule, augmentation views, auxiliary losses, and teacher update rules differently, but this notation captures the shared unlabeled self-training branch.

%

\textbf{Random-Target Supervised Mixing (RTSM).}
%
RTSM measures the direct benefit of sparse target labels before introducing any additional label-guided mechanism.
It treats labeled target images as ordinary supervised examples while preserving the original SFDA-OD self-training branch.
Let $\mathcal{D}_t^l=\{(x_i^l,y_i^l)\}_{i=1}^{N_l}$ be a uniformly random subset of target images annotated with human labels.
The remaining target images form the unlabeled set $\mathcal{D}_t^u$.
Given an SFDA-OD baseline with unlabeled objective $\mathcal{L}_{\mathrm{u}}$, RTSM optimizes
\begin{equation}
    \mathcal{L}_{\mathrm{RTSM}}
    =
    \mathcal{L}_{\mathrm{u}}
    +
    \frac{1}{|\mathcal{B}_l|}
    \sum_{(x_i^l,y_i^l) \in \mathcal{B}_l}
    \mathcal{L}_{\mathrm{det}}\!\left(f_{\theta_t}(x_i^l), y_i^l\right),
\end{equation}
where $\mathcal{B}_l$ is a labeled target mini-batch.
The pseudo-label rule, the unlabeled objective, and the teacher-student update remain those of the original SFDA-OD baseline.
RTSM therefore measures the gain from direct target supervision before any additional method attempts to use the same labels to improve unlabeled self-training.
%

%

\section{Sparse Label Feedback for Target Self-Training}
\label{sec:method}


RTSM extends a SFDA-OD baseline by adding a supervised loss on a randomly labeled target subset, while preserving the original self-training pipeline as described in Section~\ref{sec:prelim}. 
This section studies whether the same sparse annotations provide useful feedback to the unlabeled branch beyond this direct supervised loss. 
We therefore keep the RTSM as the anchor and evaluate a set of modular feedback mechanisms that intervene in different parts of target self-training.

\begin{figure}[htb]
    \centering
    \includegraphics[width=0.9\linewidth]{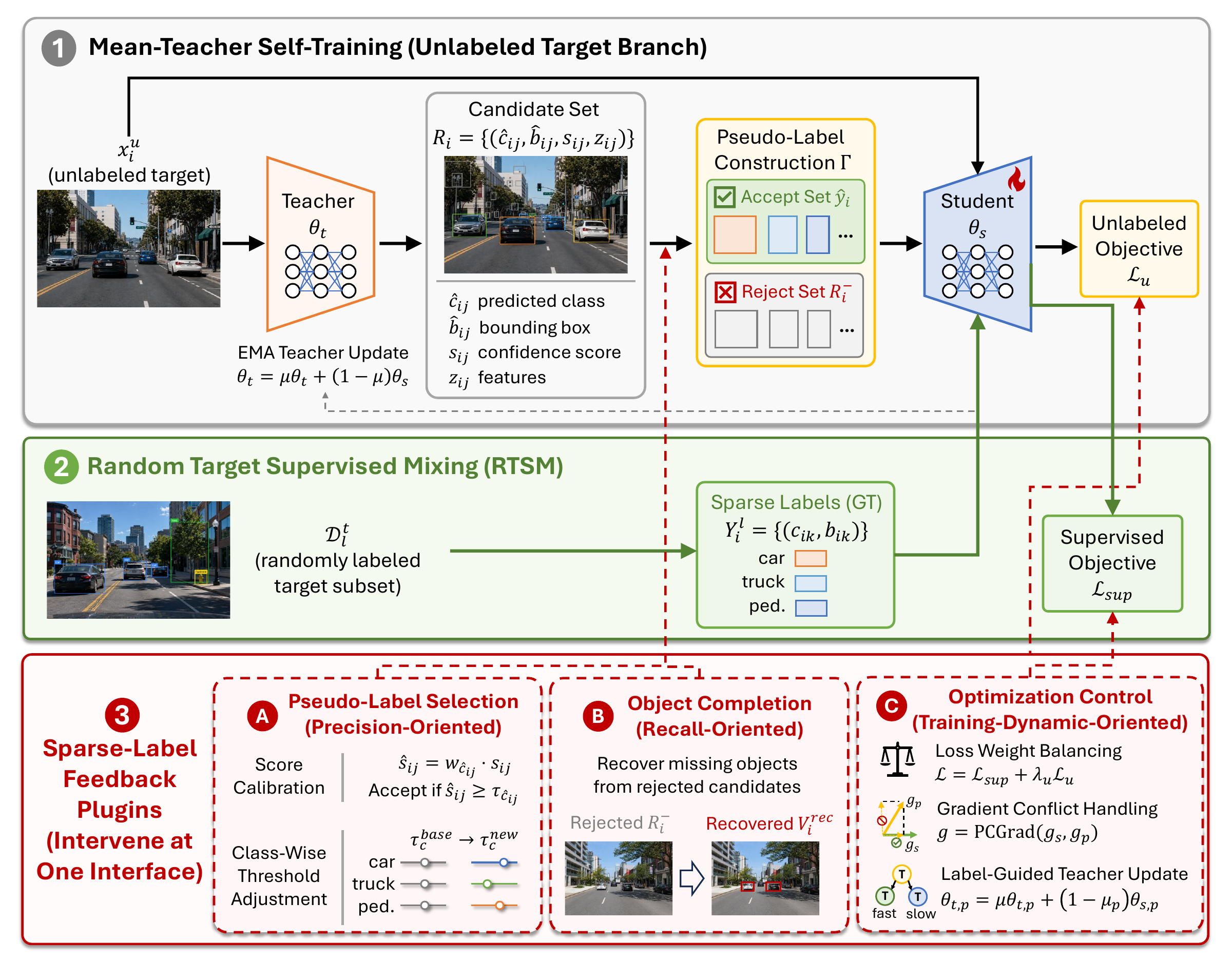}
    \vspace{-6pt}
    \caption{Sparse label feedback for target self-training. RTSM uses the randomly labeled target subset through the direct supervised detection loss. Plugin modules use the same sparse annotations as feedback to the unlabeled branch through pseudo-label selection, object completion, or optimization control.}
    \label{fig:sparse_label_feedback}
\vspace{-15pt}
\end{figure}

The mean teacher pipeline offers three natural intervention points, as illustrated in Figure~\ref{fig:sparse_label_feedback}. 
\textbf{Pseudo-Label Selection} acts before teacher predictions become pseudo-labels and tests whether sparse annotations can make confidence-based pseudo-label construction more target-aware. 
\textbf{Object Completion} acts on predictions rejected by the baseline pseudo-label construction and tests whether sparse annotations can identify target objects that the teacher localizes but fails to keep. 
\textbf{Optimization Control} leaves pseudo-labels unchanged and instead uses sparse annotations to regulate the interaction between the supervised target loss, the unlabeled objective, and the teacher update. 
In all cases, the supervised sparse target branch is unchanged, and each plugin modifies only one interface in the unlabeled branch.

\subsection{Pseudo-Label Selection}
\label{subsec:pseudo_label_selection}

Pseudo-label selection examines whether sparse annotations can improve the way teacher predictions are converted into pseudo-labels on the target domain. 
For each target image $x_i$, the teacher produces candidates $\mathcal{R}_i=\{r_{ij}\}_{j=1}^{M_i}$, where $r_{ij}=(\hat{c}_{ij},\hat{\mathbf{b}}_{ij},s_{ij},\mathbf{z}_{ij})$ contains the predicted class, box, confidence score, and candidate feature. 
The SFDA-OD baseline then applies its pseudo-label construction rule $\Gamma$ to obtain $\hat{y}_i=\Gamma(\mathcal{R}_i)$. 
Since teacher confidence can be miscalibrated after domain shift, the following three plugins keep the candidate pool fixed and change only the scores or thresholds used by $\Gamma$. 

\textbf{Prior-Mapped Thresholding.}
This plugin uses the sparse target annotations to correct class-level imbalance in the pseudo-label set. 
Its premise is that, although a small random labeled subset cannot provide exact target priors, large discrepancies between the class distribution of sparse ground truth and that of baseline pseudo-labels can still indicate systematic over-selection or under-selection. 

Let $\pi_c^{l}$ and $\pi_c^{p}$ denote the smoothed prior for class $c$ estimated from sparse ground truth objects and baseline pseudo-labels on the same sparse images. The prior discrepancy is mapped to a bounded threshold direction,
\begin{equation}
    d_c
    =
    \tanh
    \left(
    \beta\log\frac{\pi_c^{l}+\epsilon}{\pi_c^{p}+\epsilon}
    \right),
\end{equation}
where $\beta$ controls sensitivity and $\epsilon$ is used for numerical stability. 
When $\pi_c^{l}>\pi_c^{p}$, the positive value of $d_c$ lowers the threshold for class $c$ and allows more teacher candidates to enter the pseudo-label set. 
When $\pi_c^{l}<\pi_c^{p}$, the negative value of $d_c$ raises the threshold and suppresses over-represented pseudo-labels. 

Therefore given the baseline threshold $\tau_c^{0}$, the adjusted class threshold is
\begin{equation}
    \tau_c^{\mathrm{prior}}
    =
    \operatorname{clip}
    \left(
    \tau_c^{0}
    -
    \delta_{\downarrow}[d_c]_+
    +
    \delta_{\uparrow}[-d_c]_+,
    \tau_{\min},
    \tau_{\max}
    \right).
\end{equation}
Here $[x]_+=\max(x,0)$, $\delta_{\downarrow}$ and $\delta_{\uparrow}$ bound the largest downward and upward shifts, and $\tau_{\min}$ and $\tau_{\max}$ define the valid threshold range. The adjusted threshold replaces $\tau_c^{0}$ when $\Gamma$ is applied to unlabeled target images.
The prior estimation and threshold mapping details are provided in Appendix~\ref{app:selection_prior_details}.

\textbf{Precision-Calibrated Thresholding.}
Prior-mapped thresholding uses class counts but does not verify candidate correctness. 
Precision-calibrated thresholding uses sparse annotations as a small target validation set for the teacher candidates. 
For each class, it tests a grid of alternative confidence thresholds and asks which threshold retains the largest number of correct candidates while satisfying a required empirical precision.

For class $c$, let $\mathcal{T}_c$ denote the threshold grid, $P_c(\tau)$ denote the empirical precision of class $c$ candidates retained at threshold $\tau$ on sparse labeled images, and $R_c(\tau)$ denote the retained coverage of correct teacher candidates for that class. 
Subject to a target precision $p_{\min}$ and a minimum support requirement $n_{\min}$, the calibrated threshold is
\begin{equation}
    \tau_c^{\mathrm{prec}}
    =
    \arg\max_{\tau\in\mathcal{T}_c} R_c(\tau)
    \quad
    \mathrm{s.t.}
    \quad
    P_c(\tau)\geq p_{\min},
    \quad
    |\mathcal{A}_c(\tau)|\geq n_{\min},
\end{equation}
where $\mathcal{A}_c(\tau)$ is the set of sparse-image teacher candidates of class $c$ whose confidence is at least $\tau$. If the sparse subset does not provide a feasible threshold, the method falls back to the baseline threshold $\tau_c^{0}$.
The threshold grid, candidate matching, and tie-breaking rule are described in Appendix~\ref{app:selection_precision_details}.

\textbf{Reliability Score Rescaling.}
The previous two plugins replace class-specific thresholds. 
Reliability rescaling keeps the baseline thresholds intact and instead suppresses scores for classes whose accepted pseudo-labels are unreliable on sparse labeled images. 
It is therefore a conservative variant of selection, since it can remove doubtful candidates but does not explicitly increase recall.

Let $q_c=P_c(\tau_c^{0})$ be the empirical precision of class $c$ at the baseline threshold. When class $c$ satisfies the support requirements defined in Appendix~\ref{app:selection_reliability_details}, its score multiplier is
\begin{equation}
    w_c
    =
    w_{\min}
    +
    (1-w_{\min})
    \left[
    \operatorname{clip}
    \left(
    \frac{q_c}{p_{\min}},0,1
    \right)
    \right]^{\gamma}.
\end{equation}
If support is insufficient, $w_c$ is set to one so that the class is left unchanged. 
During pseudo-label construction, each teacher score is replaced by $\tilde{s}_{ij}=w_{\hat{c}_{ij}}s_{ij}$ before applying the baseline rule $\Gamma$.

\subsection{Object Completion}
\label{subsec:object_completion}
 
Object completion targets a different failure mode from pseudo-label selection. 
Selection is precision-oriented because it changes which teacher candidates are trusted. 
Completion is recall-oriented because it asks whether rejected teacher candidates can recover target objects missing from the accepted pseudo-label set. 
This distinction is important for query-based detectors, where low-confidence object queries can still localize real objects but are removed before supervising the student.

Sparse labeled images provide supervision for identifying such rejected candidates. 
On a sparse image, the baseline pseudo-label set $\hat{y}_i=\Gamma(\mathcal{R}_i)$ is compared with ground truth annotations. 
A ground truth object is considered missed if no accepted pseudo-label of the same class overlaps it above a coverage threshold. 
A teacher candidate rejected by $\Gamma$ is then labeled as a recovery positive if it matches a missed object of the same class. 
These labels train candidate scorers that are later applied to unlabeled images.
The missed-object definition and recovery-label construction are given in Appendix~\ref{app:completion_label_details}.

\textbf{Hard Query Recovery.}
Hard query recovery directly expands the pseudo-label set. 
The sparse subset is used to train a binary scorer $h_{\psi}$ that predicts whether a rejected candidate corresponds to a missed object. 
The input feature $\phi_{ij}$ includes information from the detector outputs as detailed in Appendix~\ref{app:completion_feature_details}. 
The scorer is trained with the constructed binary labels.

During domain adaptation, rejected candidates in $\mathcal{R}_i^{-} = \mathcal{R}_i \setminus \hat{y}_i$ with recovery score at least $\rho$ are appended to the baseline pseudo-label set after duplicate removal,
\begin{equation}
    \hat{y}_i^{\mathrm{rec}}
    =
    \hat{y}_i
    \cup
    \left\{
    (\hat{c}_{ij},\hat{\mathbf{b}}_{ij})
    \mid
    r_{ij}\in\mathcal{R}_i^{-},
    h_{\psi}(\phi_{ij})\geq\rho,
    \operatorname{dup}(r_{ij},\hat{y}_i)=0
    \right\}.
\end{equation}
The student is then trained on $\hat{y}_i^{\mathrm{rec}}$ in place of the baseline pseudo-label set $\hat{y}_i$.
The scorer training objective, recovery thresholding, and duplicate removal rule are described in Appendix~\ref{app:completion_hard_details}.

\textbf{Multi-View Hard Recovery.}
Multi-view hard recovery augments the same recovery scorer with cross-view stability. 
The teacher is evaluated on multiple weak augmentations of the same image, following the common consistency-training convention of producing pseudo-labels from content-preserving views~\cite{Sohn_2020_NeurIPS_FixMatch,Liu_2021_ICLR_UnbiasedTeacher}, and predictions are mapped back to the original coordinates. 
A rejected candidate receives higher support when candidates with the same predicted class and similar box locations appear in other views. This support is concatenated with the single-view feature $\phi_{ij}$ before scorer training and inference. 
The final action is unchanged from hard query recovery, namely high-scoring recovered candidates are appended to the pseudo-label set after duplicate removal.
The multi-view candidate construction and cross-view support features are detailed in Appendix~\ref{app:completion_multiview_details}.

\textbf{Foreground Query Revival.}
Hard recovery treats each selected candidate as a full pseudo-label, which can propagate errors in both class and localization. 
Foreground query revival uses the same recovered candidates as weaker supervision. It does not insert them into $\hat{y}_i$. 
Instead, it identifies the student prediction on the strong view that best corresponds to each recovered candidate and encourages that prediction to remain foreground-like.

Let $\mathcal{V}_i$ be the recovered candidate set for image $i$, let $\mu(i,j)$ denote the matched student prediction for candidate $r_{ij}$, and let $\ell_{\mu(i,j)}$ be its class logits. A foreground aggregation operator $\operatorname{fg}(\cdot)$ converts foreground-class logits into one scalar logit. The revival loss is
\begin{equation}
    \mathcal{L}_{\mathrm{rev}}
    =
    \frac{1}{|\mathcal{V}_i|+\epsilon}
    \sum_{r_{ij}\in\mathcal{V}_i}
    \operatorname{BCEWithLogits}
    \left(
    \operatorname{fg}(\ell_{\mu(i,j)}),
    1
    \right),
\end{equation}
The unlabeled objective becomes $\mathcal{L}_{\mathrm{u}}+\lambda_{\mathrm{rev}}\mathcal{L}_{\mathrm{rev}}$. 
We also evaluate \textbf{Multi-View Foreground Revival}, which constructs $\mathcal{V}_i$ using the multi-view recovery scorer and applies the same foreground revival loss.
Additional details of foreground revival are provided in Appendix~\ref{app:completion_revival_details}.

\subsection{Optimization Control}
\label{subsec:optimization_control}

Optimization control uses sparse annotations after pseudo-labels have already been constructed. 
Instead of changing candidate selection or recovering missed objects, this family treats the sparse supervised loss as a trusted target signal for controlling how the student and teacher are updated.

\textbf{Sparse Loss Balancing.}
Sparse loss balancing adjusts the strength of the unlabeled branch according to its magnitude relative to the sparse supervised loss. 
This is useful because different SFDA-OD baselines define $\mathcal{L}_{\mathrm{u}}$ differently, ranging from a hard pseudo-label detection loss to a combination of pseudo-label, consistency, distillation, and auxiliary terms. 
A fixed coefficient can therefore make the unlabeled branch either dominate sparse target supervision or contribute too weakly to adaptation.

Let $\bar{L}_{\mathrm{sup}}$ and $\bar{L}_{\mathrm{u}}$ be exponential moving averages of supervised objective $\mathcal{L}_{\mathrm{sup}}$ and self-training unlabeled objective $\mathcal{L}_{\mathrm{u}}$.  
The scale applied to the unlabeled branch is
\begin{equation}
    \lambda_t
    =
    \operatorname{clip}
    \left(
    \left(
    \frac{\kappa\bar{L}_{\mathrm{sup}}}{\bar{L}_{\mathrm{u}}+\epsilon}
    \right)^{\alpha},
    \lambda_{\min},
    \lambda_{\max}
    \right),
\end{equation}
where $\kappa$ is the desired relative magnitude, $\alpha$ controls the adaptation rate, and the clipping range prevents unstable loss scaling. The student objective becomes
\begin{equation}
    \mathcal{L}
    =
    \mathcal{L}_{\mathrm{sup}}
    +
    \lambda_t\mathcal{L}_{\mathrm{u}}.
\end{equation}

\textbf{Target-Anchored PCGrad.}
Inspired by multi-task learning~\cite{Zhang_2017_MTLSurvey}, target-anchored PCGrad asks whether the pseudo-label detection loss updates the model in directions that conflict with sparse target supervision.
Let $g_s=\nabla_{\theta}\mathcal{L}_{\mathrm{sup}}$ be the supervised sparse-label gradient and let $g_p=\nabla_{\theta}\mathcal{L}_{\mathrm{p}}$ be the gradient of the hard pseudo-label detection loss inside $\mathcal{L}_{\mathrm{u}}$.  
Following PCGrad~\cite{Yu_2020_NeurIPS_PCGrad}, when $g_p$ is negatively aligned with $g_s$, the pseudo-label update would move parameters against the trusted target supervision. We therefore project the pseudo-label gradient to remove the component that opposes the supervised gradient,
\begin{equation}
    g_p'
    =
    g_p
    -
    \mathbf{1}[g_p^{\top}g_s<0]
    \frac{g_p^{\top}g_s}{\|g_s\|_2^2+\epsilon}
    g_s.
\end{equation}
The optimizer uses the supervised gradient together with the projected pseudo-label gradient. Auxiliary unlabeled terms are left unchanged.

\textbf{Label-Guided Teacher Update.}
Instead of affecting the student training directly, sparse annotations can also be used to guide the teacher update, which affects future pseudo-label generation and student training in an indirect way. 
Specifically, the standard EMA updates all teacher parameters with a shared momentum, while sparse labels provide an additional criterion for identifying parameter groups that are important for target-domain supervision.

For a teacher parameter group $p$, let $I_l(p)$ be the normalized gradient magnitude induced by $\mathcal{L}_{\mathrm{sup}}$ on a sparse labeled batch.
Let $I(p)$ denote the merged importance score obtained from sparse supervision and any baseline-specific unlabeled importance signal, and let $\zeta_q$ be the cutoff value for the top $q_{\mathrm{imp}}$ fraction of $I(p)$.
Parameters with high merged importance receive a more target-responsive EMA momentum,
\begin{equation}
    \mu_p
    =
    \begin{cases}
    \mu_{\mathrm{fast}}, & I(p)\geq \zeta_q,\\
    \mu_{\mathrm{base}}, & \mathrm{otherwise}.
    \end{cases}
\end{equation}

The optimization control together with pseudo-label selection and object completion cover the principal ways in which sparse target labels can interact with mean teacher SFDA-OD. 
We provide the full details and hyperparameter summary in Appendix~\ref{app:full_plugin_design}.
The experiments in Section~\ref{sec:experiments} evaluate whether any of these feedback routes consistently improves over the RTSM anchor.

%
%

\section{Experiments}
\label{sec:experiments}

\subsection{Experimental Setup}
\label{subsec:experimental_setup}
%
\textbf{Datasets.}
Following existing SFDA-OD evaluations~\cite{He_2025_ICCV_DDT}, we conduct experiments on two target-shift settings constructed from three detection datasets.
\textbf{Cityscapes} contains real urban driving scenes with eight common object categories~\cite{Cordts_2016_Cityscapes}.
\textbf{Foggy Cityscapes} is derived from Cityscapes by rendering synthetic fog at multiple density levels, and we use the standard fog density of $0.02$ as the target domain~\cite{Sakaridis_2018_FoggyCityscapes}.
\textbf{BDD100K} provides real-world driving images collected under diverse scene, weather, and camera conditions~\cite{Yu_2020_BDD100K}.
Following prior SFDA-OD work, we use its daytime subset for adaptation and evaluation.
Based on these datasets, we evaluate the adaptation tasks of Cityscapes $\rightarrow$ Foggy Cityscapes for normal-to-foggy adaptation and Cityscapes $\rightarrow$ BDD100K for cross-scene adaptation.

%
\textbf{Detectors and Metrics.}
We conduct experiments with two transformer-based object detectors, \textbf{DINO}~\cite{Zhang_2022_DINO} and \textbf{DETA}~\cite{OuyangZhang_2022_DETA} with a ResNet-50 backbone.
We report \textbf{AP50} as the evaluation metric, which is the mean average precision over categories at an IoU threshold of $0.5$.
Since our experiments involve different random labeled subsets and training seeds, we report the mean and standard deviation of AP50 across runs.

%
\textbf{SFDA-OD Methods.}
We instantiate RTSM and all sparse-label feedback plugins on top of four representative SFDA-OD methods.
\begin{itemize}[noitemsep,nolistsep,topsep=0pt,leftmargin=*]
\item \textbf{PETS:} PETS~\cite{Liu_2023_ICCV_PETS} (ICCV 2023) improves teacher-student self-training by periodically exchanging the roles of teacher and student detectors.
\item \textbf{LPU:} LPU~\cite{Chen_2023_ACMMM_LPU} (ACM MM 2023) further studies the use of low-confidence predictions through proposal-level soft training and local contrastive learning.
\item \textbf{LPLD:} LPLD~\cite{Yoon_2024_ECCV_LPLD} (ECCV 2024) exploits low-confidence pseudo-labels to reduce the information loss caused by high-confidence filtering.
\item \textbf{DDT:} DDT~\cite{He_2025_ICCV_DDT} (ICCV 2025) uses a dual-rate dynamic teacher to improve source-free teacher-student adaptation.
\end{itemize}
These methods cover different pseudo-label construction and teacher-student adaptation strategies, allowing us to evaluate whether sparse-label feedback behaves consistently across SFDA-OD designs.

%
\textbf{Baselines and Compared Variants.}
Our evaluation asks whether sparse target annotations can improve unlabeled target self-training beyond their direct supervised use.
For each SFDA-OD method, the primary comparison is against RTSM, which adds a supervised detection loss on a uniformly random labeled target subset while keeping the original unlabeled branch unchanged.
We compare RTSM with ten plugin variants from Section~\ref{sec:method} and all plugin variants use the same labeled subset as RTSM.
We also include three reference baselines.
\begin{itemize}[noitemsep,nolistsep,topsep=0pt,leftmargin=*]
\item \textbf{Source-only:} evaluates the source-trained detector on the target validation set without target adaptation.
\item \textbf{Full-target oracle:} trains with all target training annotations and serves as an upper reference for the remaining target-domain headroom.
\item \textbf{Pure SFDA:} applies the original source-free adaptation method without labeled target images.
\end{itemize}

%
\textbf{Other Settings.}
The default sparse-label budget is $5\%$ of the target training images.
We evaluate $1\%$ and $10\%$ budgets in Section~\ref{sec:ablation_study}.
For each experimental run, the randomly labeled target subset is sampled with multiple seeds and is shared by RTSM and all plugin variants under the same run.
More plugin hyperparameters are reported in Appendix~\ref{app:plugin_hyperparameters}.

\subsection{Main Results}
\label{sec:main_results}

Table~\ref{tab:main_results_foggy_dino} reports the main results on Cityscapes $\rightarrow$ Foggy Cityscapes with DINO under the $5\%$ random target-label budget.
For each SFDA-OD method, we compare each plugin against its matched RTSM anchor as shown in the $\Delta$ column.

\begin{table}[htb]
    \centering
    \caption{
        Results on \textbf{Cityscapes $\rightarrow$ Foggy Cityscapes} with DINO under the 5\% random target-label budget.
        We report the mean AP50 with standard deviation.
        For plugin rows, $\Delta$ column reports the AP50 difference from the matched RTSM anchor, while Green and red cells indicate corresponding improvement or degradation.
    }
    \label{tab:main_results_foggy_dino}
    \vspace{-10pt}
    \scriptsize
    \setlength{\tabcolsep}{0.045cm}
    \renewcommand{\arraystretch}{1.08}
    \newcommand{\apstd}[2]{#1{\scriptsize$_{\pm #2}$}}
    \newcommand{\posdelta}[1]{\cellcolor[HTML]{DDF8CB}\textcolor[HTML]{1B6E1B}{+#1}}
    \newcommand{\negdelta}[1]{\cellcolor[HTML]{F8D7DA}\textcolor[HTML]{8A1C1C}{#1}}
    \resizebox{\linewidth}{!}{
    \begin{tabular}{@{}l*{4}{rc}@{}}
        \toprule
        \multirow{2}{*}{\begin{tabular}[c]{@{}l@{}}Protocol /\\ Plugin\end{tabular}} &
        \multicolumn{2}{c}{PETS} &
        \multicolumn{2}{c}{LPU} &
        \multicolumn{2}{c}{LPLD} &
        \multicolumn{2}{c}{DDT} \\
        \cmidrule(lr){2-3}\cmidrule(lr){4-5}\cmidrule(lr){6-7}\cmidrule(l){8-9}
        & AP50 & $\Delta$ & AP50 & $\Delta$ & AP50 & $\Delta$ & AP50 & $\Delta$ \\
        \midrule
        \rowcolor[HTML]{F5F5F5}
        Source-Only & \multicolumn{8}{c}{32.608} \\
        \rowcolor[HTML]{F5F5F5}
        Full-Target Oracle & \multicolumn{8}{c}{62.507} \\
        \rowcolor[HTML]{F9F9F9}
        Pure SFDA
        & \multicolumn{2}{c}{\apstd{46.520}{1.299}}
        & \multicolumn{2}{c}{\apstd{48.695}{1.266}}
        & \multicolumn{2}{c}{\apstd{47.043}{2.896}}
        & \multicolumn{2}{c}{\apstd{53.669}{0.350}} \\
        \rowcolor[HTML]{F9F9F9}
        RTSM
        & \multicolumn{2}{c}{\apstd{52.885}{0.959}}
        & \multicolumn{2}{c}{\apstd{53.420}{1.871}}
        & \multicolumn{2}{c}{\apstd{55.321}{1.558}}
        & \multicolumn{2}{c}{\apstd{58.695}{0.644}} \\
        \midrule
        \rowcolor[HTML]{D3D3D3}
        \multicolumn{9}{@{}l}{\hspace{0.4em}\textbf{Pseudo-Label Selection}} \\
        Prior-Mapped Thresholding
        & \apstd{53.467}{2.033} & \posdelta{0.582}
        & \apstd{55.076}{1.615} & \posdelta{1.656}
        & \apstd{54.957}{0.902} & \negdelta{-0.364}
        & \apstd{59.005}{1.068} & \posdelta{0.311} \\
        Precision-Calibrated Thresholding
        & \textbf{\apstd{56.611}{1.126}} & \posdelta{3.726}
        & \apstd{54.338}{3.117} & \posdelta{0.919}
        & \apstd{54.806}{0.718} & \negdelta{-0.515}
        & \apstd{58.631}{0.938} & \negdelta{-0.064} \\
        Reliability Score Rescaling
        & \apstd{55.008}{1.171} & \posdelta{2.123}
        & \textbf{\apstd{56.009}{0.784}} & \posdelta{2.590}
        & \apstd{55.182}{2.274} & \negdelta{-0.139}
        & \apstd{59.112}{0.242} & \posdelta{0.418} \\
        \midrule
        \rowcolor[HTML]{D3D3D3}
        \multicolumn{9}{@{}l}{\hspace{0.4em}\textbf{Object Completion}} \\
        Hard Query Recovery
        & \apstd{56.012}{1.638} & \posdelta{3.127}
        & \apstd{48.170}{2.510} & \negdelta{-5.249}
        & \apstd{48.385}{0.303} & \negdelta{-6.936}
        & \apstd{53.742}{0.764} & \negdelta{-4.953} \\
        Multi-View Hard Recovery
        & \apstd{50.607}{1.079} & \negdelta{-2.278}
        & \apstd{46.512}{2.403} & \negdelta{-6.907}
        & \apstd{48.188}{1.336} & \negdelta{-7.133}
        & \apstd{53.393}{1.851} & \negdelta{-5.302} \\
        Foreground Query Revival
        & \apstd{54.516}{1.197} & \posdelta{1.631}
        & \apstd{54.293}{1.812} & \posdelta{0.873}
        & \apstd{53.975}{1.045} & \negdelta{-1.346}
        & \apstd{59.012}{1.353} & \posdelta{0.317} \\
        Multi-View Foreground Revival
        & \apstd{54.678}{0.748} & \posdelta{1.793}
        & \apstd{55.759}{1.161} & \posdelta{2.339}
        & \textbf{\apstd{55.863}{0.853}} & \posdelta{0.542}
        & \apstd{58.954}{0.830} & \posdelta{0.259} \\
        \midrule
        \rowcolor[HTML]{D3D3D3}
        \multicolumn{9}{@{}l}{\hspace{0.4em}\textbf{Optimization Control}} \\
        Sparse Loss Balancing
        & \apstd{55.102}{1.221} & \posdelta{2.217}
        & \apstd{55.655}{0.926} & \posdelta{2.236}
        & \apstd{55.264}{1.077} & \negdelta{-0.057}
        & \apstd{58.820}{1.017} & \posdelta{0.125} \\
        Target-Anchored PCGrad
        & \apstd{54.953}{1.308} & \posdelta{2.068}
        & \apstd{55.229}{2.283} & \posdelta{1.810}
        & \apstd{54.463}{2.116} & \negdelta{-0.858}
        & \textbf{\apstd{59.147}{0.770}} & \posdelta{0.452} \\
        Label-Guided Teacher Update
        & \apstd{54.138}{2.108} & \posdelta{1.252}
        & \apstd{52.611}{2.725} & \negdelta{-0.808}
        & \apstd{55.247}{1.028} & \negdelta{-0.073}
        & \apstd{58.577}{1.387} & \negdelta{-0.118} \\
        \bottomrule
    \end{tabular}
    }
\vspace{-8pt}
\end{table}

\textbf{Sparse target supervision is a strong anchor.}
RTSM consistently improves all four SFDA-OD methods over their pure source-free adaptation baselines.
The gains are substantial, increasing AP50 by $6.365$ for PETS, $4.725$ for LPU, $5.026$ for DDT, and $8.278$ for LPLD.
This confirms that a small random subset of target annotations provides a strong and reliable adaptation signal when used directly as supervised target data.

\textbf{Sparse-label feedback gives inconsistent improvement over RTSM.}
The plugin results show no consistent winner across SFDA-OD methods.
The best-performing variant changes with the underlying method, with precision-calibrated thresholding performing best on PETS, reliability score rescaling on LPU, multi-view foreground revival on LPLD, and target-anchored PCGrad on DDT.
The signs of improvement are also inconsistent within each plugin family.
For example, selection plugins improve several RTSM anchors but degrade LPLD in most cases.
This method-dependent behavior indicates that sparse-label feedback does not transfer as a stable plug-in improvement across different SFDA-OD pipelines.

\textbf{Class-wise results expose the instability of sparse-label feedback.}
Table~\ref{tab:per_class_foggy_dino_ddt} shows that the gains of individual plugins are fragmented across categories rather than systematic. 
Improvements on some categories are often offset by degradation on others. 
For example, target-anchored PCGrad achieves the strongest plugin result for person and bus, while reducing AP50 for truck, Motor, and bicycle. 
Precision-calibrated thresholding improves car, bus, train, and bicycle, but degrades person, rider, truck, and Motor.  
Together with Table~\ref{tab:main_results_foggy_dino}, these results indicate that sparse labels are effective as direct supervised target examples, but do not provide a uniformly reliable signal for calibrating pseudo-labels or steering unlabeled self-training. 

Overall, \textit{Random target labels are most reliably beneficial through direct supervised loss, while their additional use for pseudo-label selection, object completion, or optimization control yields limited and method-dependent gains.}

\begin{table}[htb]
    \centering
    \caption{
        Class-wise AP50 on Cityscapes $\rightarrow$ Foggy Cityscapes with DINO and DDT under the 5\% random target-label budget.
        For plugin rows, green and red cells indicate per-class improvement and degradation relative to RTSM, respectively.
    }
    \label{tab:per_class_foggy_dino_ddt}
    \vspace{-10pt}
    \tiny
    \setlength{\tabcolsep}{4pt}
    \renewcommand{\arraystretch}{1.08}
    \newcommand{\pcpos}[1]{\cellcolor[HTML]{DDF8CB}\textcolor[HTML]{1B6E1B}{#1}}
    \newcommand{\pcneg}[1]{\cellcolor[HTML]{F8D7DA}\textcolor[HTML]{8A1C1C}{#1}}
    \begin{tabular}{@{}l*{8}{c}@{}}
        \toprule
        \multirow{2}{*}{\begin{tabular}[c]{@{}l@{}}Protocol /\\ Plugin\end{tabular}} &
        \multicolumn{8}{c}{Class-Wise AP50} \\
        \cmidrule(l){2-9}
        & Person & Rider & Car & Truck & Bus & Train &
       Motor & Bicycle \\
        \midrule
        \rowcolor[HTML]{F9F9F9}
        Pure SFDA
        & 59.255 & 65.057 & 73.441 & 32.533 & 56.956 & 41.660 & 43.097 & 58.379 \\
        \rowcolor[HTML]{F9F9F9}
        RTSM
        & 62.605 & 65.954 & 78.052 & 40.583 & 63.706 & 49.402 & 46.489 & 59.599 \\
        \midrule
        \rowcolor[HTML]{D3D3D3}
        \multicolumn{9}{@{}l}{\hspace{0.4em}\textbf{Pseudo-Label Selection}} \\
        Prior-Mapped Thresholding
        & \pcneg{62.345} & \pcneg{65.631} & \pcpos{78.282} & \pcpos{\textbf{42.229}} & \pcneg{63.170} & \pcpos{50.897} & \pcneg{46.109} & \pcneg{59.425} \\
        Precision-Calibrated Thresholding
        & \pcneg{61.739} & \pcneg{65.355} & \pcpos{78.185} & \pcneg{39.170} & \pcpos{65.055} & \pcpos{\textbf{51.832}} & \pcneg{45.115} & \pcpos{\textbf{60.126}} \\
        Reliability Score Rescaling
        & \pcneg{62.227} & \pcneg{65.470} & \pcpos{78.415} & \pcpos{41.023} & \pcneg{62.683} & \pcpos{51.271} & \pcneg{45.956} & \pcneg{59.538} \\
        \midrule
        \rowcolor[HTML]{D3D3D3}
        \multicolumn{9}{@{}l}{\hspace{0.4em}\textbf{Object Completion}} \\
        Hard Query Recovery
        & \pcneg{60.469} & \pcneg{63.937} & \pcneg{74.915} & \pcneg{40.556} & \pcneg{50.666} & \pcneg{20.429} & \pcneg{44.301} & \pcneg{59.057} \\
        Multi-View Hard Recovery
        & \pcneg{60.805} & \pcneg{64.367} & \pcneg{75.134} & \pcneg{39.860} & \pcneg{52.887} & \pcneg{17.052} & \pcneg{45.996} & \pcneg{59.078} \\
        Foreground Query Revival
        & \pcneg{62.524} & \pcpos{\textbf{66.135}} & \pcpos{78.411} & \pcpos{40.917} & \pcneg{62.543} & \pcpos{49.832} & \pcpos{46.768} & \pcneg{59.481} \\
        Multi-View Foreground Revival
        & \pcneg{62.183} & \pcneg{65.322} & \pcpos{\textbf{78.489}} & \pcneg{39.696} & \pcneg{62.737} & \pcneg{46.868} & \pcpos{\textbf{46.931}} & \pcneg{59.422} \\
        \midrule
        \rowcolor[HTML]{D3D3D3}
        \multicolumn{9}{@{}l}{\hspace{0.4em}\textbf{Optimization Control}} \\
        Sparse Loss Balancing
        & \pcneg{61.771} & \pcneg{65.489} & \pcpos{78.125} & \pcneg{40.139} & \pcneg{63.616} & \pcpos{51.239} & \pcneg{45.656} & \pcneg{59.228} \\
        Target-Anchored PCGrad
        & \pcpos{\textbf{62.660}} & \pcpos{66.080} & \pcpos{78.464} & \pcneg{39.086} & \pcpos{\textbf{65.416}} & \pcpos{50.915} & \pcneg{45.707} & \pcneg{59.204} \\
        Label-Guided Teacher Update
        & \pcneg{62.590} & \pcneg{65.479} & \pcpos{78.368} & \pcneg{39.226} & \pcpos{64.147} & \pcpos{51.326} & \pcneg{46.098} & \pcpos{59.941} \\
        \bottomrule
    \end{tabular}
    \vspace{-15pt}
\end{table}

\subsection{Ablation Study}
\label{sec:ablation_study}
%
%
%
%

To assess the robustness of our main finding, we vary the adaptation task, detector, and sparse-label budget while keeping the remaining settings fixed. These experiments examine whether the strength of direct sparse supervision and the inconsistency of label-guided feedback persist.

\textbf{Varying the Adaptation Task.}
Table~\ref{tab:task_ablation_bdd_dino} extends the evaluation to Cityscapes $\rightarrow$ BDD100K, replacing synthetic fog with heterogeneous real-world variation in scenes, weather, and camera conditions. 
RTSM again provides a substantial improvement over every pure SFDA-OD baseline, with gains ranging from $4.380$ to $18.289$ AP50. 
In contrast, no feedback plugin consistently outperforms all four RTSM anchors. 
The strongest variant changes from hard query recovery for PETS and LPLD, to multi-view hard recovery for LPU, and multi-view foreground revival for DDT. 
Several plugins also reverse their effects across methods, most notably hard recovery and precision-calibrated thresholding. 
To sum up, direct sparse supervision remains robust, whereas additional feedback is strongly coupled to the underlying SFDA-OD method.

\begin{table}[htb]
    \vspace{-15pt}
    \centering
    \caption{
        Results on Cityscapes $\rightarrow$ BDD100K with DINO under the 5\% random target-label budget.
        We report the mean AP50 with standard deviation over three seeds.
        For plugin rows, $\Delta$ reports the AP50 difference from the matched RTSM anchor.
    }
    \vspace{-10pt}
    \label{tab:task_ablation_bdd_dino}
    \scriptsize
    \setlength{\tabcolsep}{0.045cm}
    \renewcommand{\arraystretch}{1.08}
    \newcommand{\bddapstd}[2]{#1{\scriptsize$_{\pm #2}$}}
    \newcommand{\bddposdelta}[1]{\cellcolor[HTML]{DDF8CB}\textcolor[HTML]{1B6E1B}{+#1}}
    \newcommand{\bddnegdelta}[1]{\cellcolor[HTML]{F8D7DA}\textcolor[HTML]{8A1C1C}{#1}}
    \resizebox{\linewidth}{!}{
    \begin{tabular}{@{}l*{4}{rc}@{}}
        \toprule
        \multirow{2}{*}{\begin{tabular}[c]{@{}l@{}}Protocol /\\ Plugin\end{tabular}} &
        \multicolumn{2}{c}{PETS} &
        \multicolumn{2}{c}{LPU} &
        \multicolumn{2}{c}{LPLD} &
        \multicolumn{2}{c}{DDT} \\
        \cmidrule(lr){2-3}\cmidrule(lr){4-5}\cmidrule(lr){6-7}\cmidrule(l){8-9}
        & AP50 & $\Delta$ & AP50 & $\Delta$ & AP50 & $\Delta$ & AP50 & $\Delta$ \\
        \midrule
        \rowcolor[HTML]{F5F5F5}
        Source-Only & \multicolumn{8}{c}{26.225} \\
        \rowcolor[HTML]{F5F5F5}
        Full-Target Oracle & \multicolumn{8}{c}{58.003} \\
        \rowcolor[HTML]{F9F9F9}
        Pure SFDA
        & \multicolumn{2}{c}{\bddapstd{36.153}{0.179}}
        & \multicolumn{2}{c}{\bddapstd{21.067}{2.746}}
        & \multicolumn{2}{c}{\bddapstd{28.465}{1.533}}
        & \multicolumn{2}{c}{\bddapstd{34.770}{0.400}} \\
        \rowcolor[HTML]{F9F9F9}
        RTSM
        & \multicolumn{2}{c}{\bddapstd{40.533}{0.187}}
        & \multicolumn{2}{c}{\bddapstd{39.356}{0.610}}
        & \multicolumn{2}{c}{\bddapstd{38.055}{0.377}}
        & \multicolumn{2}{c}{\bddapstd{48.413}{1.699}} \\
        \midrule
        \rowcolor[HTML]{D3D3D3}
        \multicolumn{9}{@{}l}{\hspace{0.4em}\textbf{Pseudo-Label Selection}} \\
        Prior-Mapped Thresholding
        & \bddapstd{40.227}{0.080} & \bddnegdelta{-0.306}
        & \bddapstd{40.988}{1.518} & \bddposdelta{1.632}
        & \bddapstd{38.894}{1.505} & \bddposdelta{0.839}
        & \bddapstd{49.195}{2.024} & \bddposdelta{0.783} \\
        Precision-Calibrated Thresholding
        & \bddapstd{40.053}{0.291} & \bddnegdelta{-0.480}
        & \bddapstd{40.305}{0.314} & \bddposdelta{0.949}
        & \bddapstd{39.064}{0.961} & \bddposdelta{1.009}
        & \bddapstd{44.814}{0.828} & \bddnegdelta{-3.598} \\
        Reliability Score Rescaling
        & \bddapstd{40.136}{0.083} & \bddnegdelta{-0.397}
        & \bddapstd{38.585}{0.766} & \bddnegdelta{-0.772}
        & \bddapstd{37.980}{0.539} & \bddnegdelta{-0.075}
        & \bddapstd{44.015}{0.662} & \bddnegdelta{-4.398} \\
        \midrule
        \rowcolor[HTML]{D3D3D3}
        \multicolumn{9}{@{}l}{\hspace{0.4em}\textbf{Object Completion}} \\
        Hard Query Recovery
        & \textbf{\bddapstd{41.164}{0.399}} & \bddposdelta{0.631}
        & \bddapstd{41.000}{0.799} & \bddposdelta{1.644}
        & \textbf{\bddapstd{40.128}{0.603}} & \bddposdelta{2.072}
        & \bddapstd{46.628}{0.543} & \bddnegdelta{-1.784} \\
        Multi-View Hard Recovery
        & \bddapstd{26.974}{1.602} & \bddnegdelta{-13.560}
        & \textbf{\bddapstd{41.578}{0.546}} & \bddposdelta{2.221}
        & \bddapstd{37.792}{1.426} & \bddnegdelta{-0.263}
        & \bddapstd{45.834}{0.445} & \bddnegdelta{-2.578} \\
        Foreground Query Revival
        & \bddapstd{39.614}{0.094} & \bddnegdelta{-0.919}
        & \bddapstd{38.403}{0.333} & \bddnegdelta{-0.953}
        & \bddapstd{37.637}{0.122} & \bddnegdelta{-0.418}
        & \bddapstd{48.526}{1.858} & \bddposdelta{0.114} \\
        Multi-View Foreground Revival
        & \bddapstd{39.828}{0.227} & \bddnegdelta{-0.705}
        & \bddapstd{39.232}{0.407} & \bddnegdelta{-0.125}
        & \bddapstd{38.068}{0.536} & \bddposdelta{0.013}
        & \textbf{\bddapstd{50.462}{1.004}} & \bddposdelta{2.049} \\
        \midrule
        \rowcolor[HTML]{D3D3D3}
        \multicolumn{9}{@{}l}{\hspace{0.4em}\textbf{Optimization Control}} \\
        Sparse Loss Balancing
        & \bddapstd{40.324}{0.513} & \bddnegdelta{-0.209}
        & \bddapstd{38.724}{0.543} & \bddnegdelta{-0.632}
        & \bddapstd{36.678}{0.938} & \bddnegdelta{-1.378}
        & \bddapstd{49.213}{0.499} & \bddposdelta{0.800} \\
        Target-Anchored PCGrad
        & \bddapstd{40.287}{0.418} & \bddnegdelta{-0.246}
        & \bddapstd{39.743}{1.110} & \bddposdelta{0.387}
        & \bddapstd{38.910}{0.837} & \bddposdelta{0.855}
        & \bddapstd{50.252}{0.124} & \bddposdelta{1.840} \\
        Label-Guided Teacher Update
        & \bddapstd{40.140}{0.261} & \bddnegdelta{-0.394}
        & \bddapstd{38.679}{0.807} & \bddnegdelta{-0.677}
        & \bddapstd{38.000}{1.203} & \bddnegdelta{-0.055}
        & \bddapstd{49.788}{1.362} & \bddposdelta{1.375} \\
        \bottomrule
    \end{tabular}
    }
    \vspace{-15pt}
\end{table}

\textbf{Varying the Detector.}
Table~\ref{tab:detector_ablation_foggy_deta} replaces DINO with DETA under the $5\%$ random target-label budget. 
RTSM remains a strong reference across SFDA-OD methods, improving AP50 over pure SFDA by $2.464$ to $10.380$ AP50. 
The plugin results, however, remain method-dependent. 
For instance, precision-calibrated thresholding improves PETS by $3.494$ AP50 but offers little benefit elsewhere, while hard recovery improves PETS yet severely degrades LPU, LPLD, and DDT. 
Multi-view foreground revival is more stable and improves all four methods, although only by $0.168$ to $0.818$ AP50. 
Thus, changing the detector preserves the central conclusion: sparse annotations transfer reliably through direct supervision, but not through a universally effective feedback mechanism.

\begin{table}[htb]
    \centering
    \caption{
        Results on Cityscapes $\rightarrow$ Foggy Cityscapes with the \textbf{DETA} detector under the 5\% random target-label budget.
        We report the mean AP50 with standard deviation.
        For plugin rows, $\Delta$ reports the AP50 difference from the matched RTSM anchor.
    }
    \label{tab:detector_ablation_foggy_deta}
    \vspace{-10pt}
    \scriptsize
    \setlength{\tabcolsep}{0.045cm}
    \renewcommand{\arraystretch}{1.08}
    \newcommand{\detaapstd}[2]{#1{\scriptsize$_{\pm #2}$}}
    \newcommand{\detaposdelta}[1]{\cellcolor[HTML]{DDF8CB}\textcolor[HTML]{1B6E1B}{+#1}}
    \newcommand{\detanegdelta}[1]{\cellcolor[HTML]{F8D7DA}\textcolor[HTML]{8A1C1C}{#1}}
    \resizebox{\linewidth}{!}{
    \begin{tabular}{@{}l*{4}{rc}@{}}
        \toprule
        \multirow{2}{*}{\begin{tabular}[c]{@{}l@{}}Protocol /\\ Plugin\end{tabular}} &
        \multicolumn{2}{c}{PETS} &
        \multicolumn{2}{c}{LPU} &
        \multicolumn{2}{c}{LPLD} &
        \multicolumn{2}{c}{DDT} \\
        \cmidrule(lr){2-3}\cmidrule(lr){4-5}\cmidrule(lr){6-7}\cmidrule(l){8-9}
        & AP50 & $\Delta$ & AP50 & $\Delta$ & AP50 & $\Delta$ & AP50 & $\Delta$ \\
        \midrule
        \rowcolor[HTML]{F5F5F5}
        Source-Only & \multicolumn{8}{c}{28.506} \\
        \rowcolor[HTML]{F5F5F5}
        Full-Target Oracle & \multicolumn{8}{c}{61.333} \\
        \rowcolor[HTML]{F9F9F9}
        Pure SFDA
        & \multicolumn{2}{c}{\detaapstd{42.660}{1.129}}
        & \multicolumn{2}{c}{\detaapstd{47.716}{2.650}}
        & \multicolumn{2}{c}{\detaapstd{48.504}{3.179}}
        & \multicolumn{2}{c}{\detaapstd{55.526}{0.524}} \\
        \rowcolor[HTML]{F9F9F9}
        RTSM
        & \multicolumn{2}{c}{\detaapstd{53.040}{1.500}}
        & \multicolumn{2}{c}{\detaapstd{55.001}{0.802}}
        & \multicolumn{2}{c}{\detaapstd{55.838}{0.796}}
        & \multicolumn{2}{c}{\detaapstd{57.990}{0.326}} \\
        \midrule
        \rowcolor[HTML]{D3D3D3}
        \multicolumn{9}{@{}l}{\hspace{0.4em}\textbf{Pseudo-Label Selection}} \\
        Prior-Mapped Thresholding
        & \detaapstd{53.491}{1.754} & \detaposdelta{0.451}
        & \detaapstd{54.991}{1.086} & \detanegdelta{-0.010}
        & \detaapstd{55.494}{0.822} & \detanegdelta{-0.344}
        & \detaapstd{58.725}{0.566} & \detaposdelta{0.735} \\
        Precision-Calibrated Thresholding
        & \textbf{\detaapstd{56.535}{0.656}} & \detaposdelta{3.494}
        & \detaapstd{55.049}{0.551} & \detaposdelta{0.048}
        & \detaapstd{55.634}{0.865} & \detanegdelta{-0.204}
        & \detaapstd{58.225}{0.328} & \detaposdelta{0.235} \\
        Reliability Score Rescaling
        & \detaapstd{53.229}{1.105} & \detaposdelta{0.188}
        & \detaapstd{55.018}{0.667} & \detaposdelta{0.017}
        & \detaapstd{55.886}{0.502} & \detaposdelta{0.048}
        & \detaapstd{58.759}{0.299} & \detaposdelta{0.769} \\
        \midrule
        \rowcolor[HTML]{D3D3D3}
        \multicolumn{9}{@{}l}{\hspace{0.4em}\textbf{Object Completion}} \\
        Hard Query Recovery
        & \detaapstd{54.893}{1.046} & \detaposdelta{1.852}
        & \detaapstd{46.345}{2.454} & \detanegdelta{-8.656}
        & \detaapstd{46.297}{3.240} & \detanegdelta{-9.542}
        & \detaapstd{52.468}{2.565} & \detanegdelta{-5.522} \\
        Multi-View Hard Recovery
        & \detaapstd{45.546}{2.675} & \detanegdelta{-7.495}
        & \detaapstd{44.140}{2.117} & \detanegdelta{-10.861}
        & \detaapstd{44.983}{2.142} & \detanegdelta{-10.856}
        & \detaapstd{50.700}{2.211} & \detanegdelta{-7.290} \\
        Foreground Query Revival
        & \detaapstd{53.622}{1.323} & \detaposdelta{0.582}
        & \detaapstd{55.177}{1.844} & \detaposdelta{0.176}
        & \detaapstd{55.259}{0.319} & \detanegdelta{-0.580}
        & \detaapstd{58.617}{0.586} & \detaposdelta{0.627} \\
        Multi-View Foreground Revival
        & \detaapstd{53.209}{0.337} & \detaposdelta{0.168}
        & \textbf{\detaapstd{55.819}{1.192}} & \detaposdelta{0.818}
        & \textbf{\detaapstd{56.136}{0.283}} & \detaposdelta{0.297}
        & \detaapstd{58.582}{0.650} & \detaposdelta{0.592} \\
        \midrule
        \rowcolor[HTML]{D3D3D3}
        \multicolumn{9}{@{}l}{\hspace{0.4em}\textbf{Optimization Control}} \\
        Sparse Loss Balancing
        & \detaapstd{53.635}{1.319} & \detaposdelta{0.595}
        & \detaapstd{55.366}{0.957} & \detaposdelta{0.365}
        & \detaapstd{55.509}{1.173} & \detanegdelta{-0.329}
        & \detaapstd{58.445}{0.365} & \detaposdelta{0.455} \\
        Target-Anchored PCGrad
        & \detaapstd{53.649}{1.327} & \detaposdelta{0.609}
        & \detaapstd{55.431}{0.440} & \detaposdelta{0.430}
        & \detaapstd{55.156}{1.145} & \detanegdelta{-0.683}
        & \detaapstd{58.618}{1.042} & \detaposdelta{0.629} \\
        Label-Guided Teacher Update
        & \detaapstd{53.433}{1.351} & \detaposdelta{0.393}
        & \detaapstd{53.623}{2.203} & \detanegdelta{-1.378}
        & \detaapstd{54.592}{1.910} & \detanegdelta{-1.247}
        & \textbf{\detaapstd{59.043}{0.810}} & \detaposdelta{1.053} \\
        \bottomrule
    \end{tabular}
    }
\vspace{-15pt}
\end{table}

\textbf{Varying the Sparse-Label Budget.}
Table~\ref{tab:budget_ablation_foggy_dino_ddt} in Appendix~\ref{app:add-exp} evaluates DDT with DINO under $1\%$, $5\%$, and $10\%$ random target-label budgets. 
Increasing the budget steadily strengthens RTSM, from $55.415$ AP50 at $1\%$ to $58.695$ at $5\%$ and $59.910$ at $10\%$. 
Nevertheless, the conclusion drawn from the main results remains unchanged. 
No plugin performs consistently across budgets, and several variants switch from positive gains at $1\%$ to neutral or negative effects at larger budgets. 

A notable secondary trend is the diminishing marginal value of plugin feedback as more target annotations become available. 
The average plugin difference from RTSM decreases from $+0.097$ AP50 at $1\%$ to $-0.856$ and $-0.816$ at $5\%$ and $10\%$, respectively. 
Under an extremely limited budget, the supervised anchor remains weak enough that feedback through the unlabeled branch can occasionally provide complementary information. 
As the labeled subset grows, direct supervision captures a larger fraction of the available target-domain signal, leaving less room for noisier and less stable interventions. 

\section{Conclusion}
\label{sec:conclusion}

This paper systematically examines how sparse target annotations should be used in source-free domain adaptive object detection. 
We establish RTSM as a simple anchor that introduces direct target supervision while preserving the original adaptation pipeline.
Across four SFDA-OD methods, two detectors, multiple domain shifts, and different label budgets, RTSM consistently improves pure source-free adaptation. 
In contrast, plugins for pseudo-label selection, object completion, and optimization control yield gains that remain method- and setting-dependent. 
These results reveal a bitter lesson: sparse labels are highly effective as direct supervision, but substantially less reliable as control signals for noisy self-training. 
RTSM should therefore serve as a necessary benchmark for future methods in this setting.
Our study focuses on transformer-based object detection and future work may investigate whether the same pattern extends to other detector families and vision tasks.

%
%
\bibliographystyle{splncs04}
\bibliography{main}

\clearpage
\newpage
\appendix
\setcounter{section}{0}

\section{Additional Related Work} \label{app:related}

Our study lies at the intersection of domain adaptive object detection, sparse-label target adaptation, and semi-supervised pseudo-label self-training.

\textbf{Source-Free Domain Adaptive Object Detection.}
Domain adaptive object detection studies how to transfer a detector trained on a labeled source domain to a target domain whose visual distribution differs from the source. 
Early and widely used DAOD methods assume access to source images during adaptation, and reduce the domain gap through image-level or instance-level feature alignment, adversarial training, asymmetric adaptation, scale-aware adaptation, or pseudo-label refinement \cite{Chen_2018_CVPR,Saito_2019_CVPR,He_2020_ATF,Shi_2022_UniDAOD}. 
However, the source-available assumption is often restrictive in practice because source data may be inaccessible due to storage, privacy, or licensing constraints.

Source-free domain adaptive object detection removes source images from the adaptation stage and relies on a source-trained detector and unlabeled target images. 
Recent SFDA-OD methods are largely built around self-training. 
A teacher detector generates pseudo-labels on target images, while a student detector is optimized using those pseudo-labels and is often stabilized through Mean Teacher updates \cite{Tarvainen_2017_NeurIPS}. 
Existing methods improve this framework from different angles. SED \cite{Li_2021_AAAI_FreeLunch} studies source-free adaptation without source data. 
PETS \cite{Liu_2023_ICCV_PETS} periodically exchanges teacher and student roles to improve self-training dynamics. 
LPU \cite{Chen_2023_ACMMM_LPU} and LPLD \cite{Yoon_2024_ECCV_LPLD} exploit low-confidence predictions to reduce the information loss caused by high-confidence pseudo-label filtering. 
DDT \cite{He_2025_ICCV_DDT} introduces a dynamic teacher update to better balance target adaptation and teacher stability. 

These methods improve the unlabeled self-training branch, while our work asks a complementary question. 
When a small random subset of target images is labeled, the direct supervised use of these labels already provides a strong anchor, and stronger sparse-label methods should demonstrate gains beyond this anchor rather than only over pure SFDA.

\textbf{Active and Sparse-Label Domain Adaptation.}
Active domain adaptation assumes an annotation budget and studies which target samples should be labeled to maximize adaptation performance. 
Existing work has explored uncertainty, diversity, domainness, representativeness, and task-specific criteria for selecting informative target images \cite{Prabhu_2020_CLUE,Wang_2022_ASFDA,Lyu_2024_LFTL}. 
In object detection, active selection is more challenging than in classification because the utility of an image depends on object categories, localization quality, missing objects, and the reliability of predicted boxes \cite{Nakamura_2024_CVPR_ADA_FN}. 
Recent active DAOD methods therefore incorporate detection-specific cues, including false-negative prediction and box-level uncertainty, to account for objects that are not captured by the current detector \cite{Nakamura_2024_CVPR_ADA_FN,Menke_2024_ESWA_ADAOD}.

Our setting is related to active and sparse-label domain adaptation, but it isolates a different question. 
Instead of optimizing the acquisition rule, we label a uniformly random subset of target images and focus on how these labels should be used after annotation. 
This protocol removes the confounding effect of sample-selection heuristics and exposes a minimal sparse-label baseline. 
The resulting RTSM anchor measures the benefit of simply mixing random target labels into the supervised detection loss of an existing SFDA-OD method. 
Our study then evaluates whether the same labels can provide additional feedback to the unlabeled branch. 

\textbf{Semi-Supervised Object Detection and Pseudo-Label Self-Training.}
Semi-supervised object detection provide the broader self-training background for our study. 
FixMatch and related methods first combine weak and strong augmentations with confidence-based pseudo-labeling \cite{Sohn_2020_NeurIPS_FixMatch}. 
Then in object detection, methods such as Unbiased Teacher, Unbiased Teacher v2, Soft Teacher, Dense Teacher, and related teacher-student frameworks address pseudo-label noise, class imbalance, and localization uncertainty in semi-supervised training \cite{Liu_2021_ICLR_UnbiasedTeacher,Liu_2022_UnbiasedTeacherV2,Xu_2021_SoftTeacher,Zhou_2022_DenseTeacher}. 

These works motivate the three feedback routes examined in this paper. 
Pseudo-label selection is related to confidence calibration and thresholding, since teacher confidence may not reflect target-domain correctness under distribution shift. 
Object completion is related to low-confidence prediction usage and missed-object recovery, since discarded candidates may contain objects that are useful for target self-training. 
Optimization control is related to loss balancing and gradient-conflict handling, since trusted labels and noisy pseudo-labels may produce different training signals. 

However, standard semi-supervised detection typically assumes that labeled and unlabeled images are sampled from the same target distribution. 
Our setting starts from a source-trained detector under target-domain shift and evaluates whether sparse target labels provide useful feedback beyond the direct RTSM supervised anchor.

\section{Full Plugin Design Details}
\label{app:full_plugin_design}

This section provides the full details for all plugin designs.
As mentioned in Section~\ref{sec:method}, we build ten modular plugins on top of the RTSM anchor to investigate the sparse-label feedback for the target self-training branch. 
The sparse labeled target subset is used by the supervised detection loss in the main training objective, while the plugins reuse the same annotations to modify one component of the unlabeled self-training branch. 
Specifically, pseudo-label selection changes how teacher candidates are retained. 
Object completion revisits candidates rejected by the baseline pseudo-label generator. 
Optimization control keeps the pseudo-label set fixed and changes the student or teacher update. 
This taxonomy ensures that each plugin tests a specific route through which sparse labels could influence unlabeled target adaptation.

\subsection{Shared Setup and Notation}
\label{app:plugin_common}

All plugins operate on the same target split. 
Let $\mathcal{D}_{l}^{t}$ denote the sparse labeled target subset and $\mathcal{D}_{u}^{t}$ denote the remaining unlabeled target subset. 
For a target image $x_i$, the teacher detector produces a candidate set $\mathcal{R}_i=\{r_{ij}\}_{j=1}^{M_i}$. 
Each candidate is represented as $r_{ij}=(\hat{c}_{ij},\hat{\mathbf{b}}_{ij},s_{ij},\mathbf{z}_{ij})$, where $\hat{c}_{ij}$ is the predicted class, $\hat{\mathbf{b}}_{ij}$ is the predicted box, $s_{ij}$ is the confidence score, and $\mathbf{z}_{ij}$ is an optional candidate-level feature. 
For query-based detectors, $\mathbf{z}_{ij}$ is the object-query feature. 
For proposal-based detectors, it can be the proposal or RoI feature. 
The baseline pseudo-label construction rule is denoted by $\Gamma$, which maps teacher candidates to the accepted pseudo-label set $\hat{y}_i=\Gamma(\mathcal{R}_i)$. The rejected candidate set is $\mathcal{R}_i^{-}=\mathcal{R}_i\setminus\hat{y}_i$.

Sparse annotations make it possible to verify teacher candidates on target-domain images. 
For $x_i\in\mathcal{D}_{l}^{t}$, let $\mathcal{G}_i=\{g_{ik}\}_{k=1}^{N_i}$ denote the ground truth object set, with $g_{ik}=(c_{ik},\mathbf{b}_{ik})$. 
A candidate $r_{ij}$ is considered correct at IoU threshold $\eta$ if it can be assigned to an unmatched ground truth object of the same class whose box overlap is at least $\eta$. When a one-to-one assignment is needed, candidates are processed in decreasing confidence order and each ground truth object can be matched at most once. We denote the resulting correctness indicator by $m_{ij}(\eta)\in\{0,1\}$. This verification procedure is used to estimate pseudo-label precision in the selection plugins and to construct recovery labels in the completion plugins.
The full notations are summarized in Table~\ref{tab:plugin_notation}.

\begin{table}[t]
\caption{Basic notation used by the sparse-label feedback plugins.}
\label{tab:plugin_notation}
\centering
\small
\setlength{\tabcolsep}{6pt}
\renewcommand{\arraystretch}{1.15}
\begin{tabular}{p{0.18\textwidth}p{0.74\textwidth}}
\hline
Symbol & Meaning \\
\hline
$\mathcal{D}_{l}^{t}$ & Sparse labeled target subset \\
$\mathcal{D}_{u}^{t}$ & Unlabeled target subset \\
$\mathcal{R}_i$ & Teacher candidate set for image $x_i$ \\
$r_{ij}$ & Candidate $j$ in image $x_i$ \\
$\hat{c}_{ij}$ & Predicted class of $r_{ij}$ \\
$\hat{\mathbf{b}}_{ij}$ & Predicted bounding box of $r_{ij}$ \\
$s_{ij}$ & Confidence score of $r_{ij}$ \\
$\mathbf{z}_{ij}$ & Optional candidate-level feature \\
$\Gamma$ & Baseline pseudo-label construction rule \\
$\hat{y}_i$ & Pseudo-label set accepted by $\Gamma$ \\
$\mathcal{R}_i^{-}$ & Teacher candidates rejected by $\Gamma$ \\
$\mathcal{G}_i$ & Ground truth object set for sparse labeled image $x_i$ \\
$m_{ij}(\eta)$ & Correctness indicator for candidate $r_{ij}$ under IoU threshold $\eta$ \\
\hline
\end{tabular}
\end{table}

\subsection{Pseudo-Label Selection Details}
\label{app:selection_details}

Pseudo-label selection modifies the candidate acceptance step of self-training. The teacher predictions themselves are unchanged. 
Each plugin either replaces the class-specific confidence thresholds used by $\Gamma$ or rescales the confidence scores before the original thresholds are applied. 
This family evaluates whether sparse target labels can provide a reliable precision-oriented signal for deciding which teacher predictions should supervise the student.

For class $c$, let $\tau_c^{0}$ denote the baseline confidence threshold. For any tested threshold $\tau$, we define the retained class-$c$ candidates on sparse images as
\begin{equation}
    \mathcal{A}_c(\tau)
    =
    \left\{
    (i,j)
    \middle|
    x_i\in\mathcal{D}_{l}^{t},
    \hat{c}_{ij}=c,
    s_{ij}\geq\tau
    \right\}.
\end{equation}
Using $m_{ij}(\eta_{\mathrm{sel}})$ to verify candidate correctness, the empirical precision at threshold $\tau$ is
\begin{equation}
    P_c(\tau)
    =
    \frac{\sum_{(i,j)\in\mathcal{A}_c(\tau)}m_{ij}(\eta_{\mathrm{sel}})}
    {|\mathcal{A}_c(\tau)|+\epsilon}.
\end{equation}
We also define the retained correct-candidate coverage as
\begin{equation}
    R_c(\tau)
    =
    \frac{\sum_{(i,j)\in\mathcal{A}_c(\tau)}m_{ij}(\eta_{\mathrm{sel}})}
    {\sum_{(i,j):x_i\in\mathcal{D}_{l}^{t},\hat{c}_{ij}=c}m_{ij}(\eta_{\mathrm{sel}})+\epsilon}.
\end{equation}
$P_c(\tau)$ measures how many retained candidates are correct. $R_c(\tau)$ measures how many correct teacher candidates of class $c$ remain after thresholding. The denominator of $R_c(\tau)$ is defined over teacher candidates rather than ground truth objects, so this quantity evaluates threshold retention rather than object recall.

\textbf{Prior-Mapped Thresholding.}
\label{app:selection_prior_details}
Prior-mapped thresholding uses the sparse labeled subset to correct class-level imbalance in pseudo-label production. The method compares the class distribution of sparse ground truth objects with the class distribution of baseline pseudo-labels on the same images. A class that appears less frequently in pseudo-labels than in sparse annotations is treated as under-produced and receives a lower threshold. A class that appears more frequently in pseudo-labels is treated as over-produced and receives a higher threshold.

Let $n_c^l$ be the number of sparse ground truth objects of class $c$, and let $n_c^p$ be the number of baseline pseudo-labels of class $c$ on the sparse images. With smoothing constant $\alpha_{\pi}$ and number of classes $C$, the class priors are
\begin{equation}
    \pi_c^l
    =
    \frac{n_c^l+\alpha_{\pi}}
    {\sum_{c'}n_{c'}^l+C\alpha_{\pi}},
    \qquad
    \pi_c^p
    =
    \frac{n_c^p+\alpha_{\pi}}
    {\sum_{c'}n_{c'}^p+C\alpha_{\pi}}.
\end{equation}
The log prior ratio is mapped to a bounded adjustment direction,
\begin{equation}
    d_c
    =
    \tanh
    \left(
    \beta\log\frac{\pi_c^{l}+\epsilon}{\pi_c^{p}+\epsilon}
    \right),
\end{equation}
where $\beta$ controls the sensitivity of the mapping and $\epsilon$ prevents numerical instability. Positive values of $d_c$ indicate that class $c$ is under-produced by the baseline pseudo-labels, while negative values indicate over-production.

The adjusted threshold is
\begin{equation}
    \tau_c^{\mathrm{prior}}
    =
    \operatorname{clip}
    \left(
    \tau_c^{0}
    -
    \delta_{\downarrow}[d_c]_+
    +
    \delta_{\uparrow}[-d_c]_+,
    \tau_{\min},
    \tau_{\max}
    \right),
\end{equation}
where $[x]_+=\max(x,0)$. The constants $\delta_{\downarrow}$ and $\delta_{\uparrow}$ bound the largest threshold decrease and increase, and $[\tau_{\min},\tau_{\max}]$ defines the valid threshold range. During unlabeled training, $\Gamma$ uses $\tau_c^{\mathrm{prior}}$ instead of $\tau_c^0$ for class $c$.

\textbf{Precision-Calibrated Thresholding.}
\label{app:selection_precision_details}
Precision-calibrated thresholding uses sparse annotations to evaluate teacher candidates directly. It searches for a class-specific threshold that retains as many correct teacher candidates as possible while meeting a target precision on sparse labeled images. This plugin therefore uses object-level verification rather than class-prior mismatch.

For each class $c$, let $\mathcal{T}_c$ be the set of tested thresholds around $\tau_c^0$. A threshold is feasible if it satisfies both a precision requirement and a minimum support requirement,
\begin{equation}
    \mathcal{F}_c
    =
    \left\{
    \tau\in\mathcal{T}_c
    \middle|
    P_c(\tau)\geq p_{\min},
    \quad
    |\mathcal{A}_c(\tau)|\geq n_{\min}
    \right\}.
\end{equation}
Here $p_{\min}$ is the required empirical precision, and $n_{\min}$ is the minimum number of retained candidates required for the estimate to be used. When $\mathcal{F}_c$ is nonempty, the selected threshold maximizes retained correct-candidate coverage,
\begin{equation}
    \tau_c^{\mathrm{prec}}
    =
    \arg\max_{\tau\in\mathcal{F}_c} R_c(\tau).
\end{equation}
If several thresholds achieve the same coverage, the implementation selects the one closest to $\tau_c^0$. If no feasible threshold exists, the class falls back to $\tau_c^0$. The resulting threshold $\tau_c^{\mathrm{prec}}$ replaces the baseline threshold when pseudo-labels are constructed on unlabeled target images.

\textbf{Reliability Rescaling.}
\label{app:selection_reliability_details}
Reliability rescaling preserves the baseline thresholds but changes the confidence scores that enter $\Gamma$. The method estimates how reliable each class is at its baseline threshold and suppresses candidate scores for classes whose accepted pseudo-labels are unreliable. Since the maximum multiplier is one, this plugin can remove doubtful predictions but cannot increase the number of accepted candidates.

The class reliability estimate is the empirical precision at the baseline threshold, $q_c=P_c(\tau_c^0)$. This estimate is used only when class $c$ has sufficient sparse evidence,
\begin{equation}
    |\mathcal{A}_c(\tau_c^{0})|\geq n_{\min},
    \qquad
    \sum_{(i,j)\in\mathcal{A}_c(\tau_c^{0})}m_{ij}(\eta_{\mathrm{sel}})\geq m_{\min}.
\end{equation}
The first condition requires enough retained candidates, and the second requires enough verified correct candidates. If either condition fails, the class multiplier is set to $w_c=1$, which leaves the baseline scores unchanged.

When the support conditions hold, the multiplier is
\begin{equation}
    w_c
    =
    w_{\min}
    +
    (1-w_{\min})
    \left[
    \operatorname{clip}
    \left(
    \frac{q_c}{p_{\min}},0,1
    \right)
    \right]^{\gamma}.
\end{equation}
$w_{\min}$ is the smallest allowed score multiplier, and $\gamma$ controls how sharply the multiplier decreases when $q_c$ falls below $p_{\min}$. For a candidate $r_{ij}$, the score used by $\Gamma$ becomes $\tilde{s}_{ij}=w_{\hat{c}_{ij}}s_{ij}$.

\subsection{Object Completion Details}
\label{app:completion_details}

Object completion addresses a recall-oriented failure mode of pseudo-label self-training. A teacher candidate may correspond to a real target object but remain below the confidence threshold, so the student never receives supervision for that object. Completion plugins use sparse annotations to learn from such rejected candidates and then apply the learned recovery rule to unlabeled target images.

\textbf{Missed Objects and Recovery Labels.}
\label{app:completion_label_details}
The recovery signal is defined by comparing accepted pseudo-labels with sparse ground truth objects. For a sparse image $x_i$, a ground truth object $g_{ik}=(c_{ik},\mathbf{b}_{ik})$ is considered covered if at least one accepted pseudo-label of the same class overlaps it above $\eta_{\mathrm{cov}}$,
\begin{equation}
    \operatorname{covered}(g_{ik})
    =
    \mathbf{1}
    \left[
    \max_{(\hat{c},\hat{\mathbf{b}})\in\hat{y}_i:\hat{c}=c_{ik}}
    \operatorname{IoU}(\hat{\mathbf{b}},\mathbf{b}_{ik})
    \geq
    \eta_{\mathrm{cov}}
    \right].
\end{equation}
The maximum over an empty set is zero. Objects not covered by accepted pseudo-labels form the missed-object set,
\begin{equation}
    \mathcal{M}_i
    =
    \left\{
    g_{ik}\in\mathcal{G}_i
    \middle|
    \operatorname{covered}(g_{ik})=0
    \right\}.
\end{equation}

A rejected candidate $r_{ij} \in \mathcal{R}_i^{-}$ receives a positive recovery label if it matches one of these missed objects with the same class and IoU at least $\eta_{\mathrm{rec}}$,
\begin{equation}
    y_{ij}^{\mathrm{rec}}
    =
    \mathbf{1}
    \left[
    \max_{g_{ik}\in\mathcal{M}_i:c_{ik}=\hat{c}_{ij}}
    \operatorname{IoU}(\hat{\mathbf{b}}_{ij},\mathbf{b}_{ik})
    \geq
    \eta_{\mathrm{rec}}
    \right].
\end{equation}
Candidates that duplicate already covered objects are treated as negatives or ignored according to the duplicate policy. This construction focuses the recovery task on teacher predictions that could add missing supervision rather than redundant detections.

\textbf{Candidate Features.}
\label{app:completion_feature_details}
Each rejected candidate is represented by a feature vector $\phi_{ij}$. The default vector includes the teacher confidence score $s_{ij}$, the margin $s_{ij}-\tau_{\hat{c}_{ij}}^0$ to the class threshold, class identity $\hat{c}_{ij}$, predictive entropy and margin uncertainty measures~\cite{Ren_2020_DALSurvey}, and normalized box center and box area. When the detector exposes candidate-level representations, the feature $\mathbf{z}_{ij}$ is appended. The same feature map is used for scorer training on sparse images and for scorer inference on unlabeled images.

\textbf{Hard Query Recovery.}
\label{app:completion_hard_details}
Hard query recovery inserts selected rejected candidates into the pseudo-label set. A binary scorer $h_{\psi}$ is trained on sparse images to predict $y_{ij}^{\mathrm{rec}}$ from $\phi_{ij}$. The training set is
\begin{equation}
    \mathcal{S}_{\mathrm{rec}}
    =
    \left\{
    (\phi_{ij},y_{ij}^{\mathrm{rec}})
    \middle|
    x_i\in\mathcal{D}_{l}^{t},
    r_{ij}\in\mathcal{R}_i^{-}
    \right\}.
\end{equation}
The scorer is optimized with the standard binary cross entropy,
\begin{equation}
    \mathcal{L}_{\mathrm{rec}}
    =
    \frac{1}{|\mathcal{S}_{\mathrm{rec}}|}
    \sum_{(\phi,y)\in\mathcal{S}_{\mathrm{rec}}}
    \operatorname{BCE}
    \left(h_{\psi}(\phi),y\right).
\end{equation}

On an unlabeled image, the baseline first obtains $\hat{y}_i=\Gamma(\mathcal{R}_i)$. The trained scorer is then applied to $\mathcal{R}_i^{-}$. A rejected candidate is inserted when $h_{\psi}(\phi_{ij})\geq\rho$ and it is not a duplicate of an accepted pseudo-label under the same-class IoU threshold $\eta_{\mathrm{dup}}$. Let $\mathcal{V}_i^{\mathrm{rec}}$ denote the recovered candidate set,
\begin{equation}
\begin{aligned}
    \mathcal{V}_i^{\mathrm{rec}}
    =
    \{(\hat{c}_{ij},\hat{\mathbf{b}}_{ij})
    \mid
    & r_{ij}\in\mathcal{R}_i^{-},
    h_{\psi}(\phi_{ij})\geq\rho, \\
    & \operatorname{dup}(r_{ij},\hat{y}_i)=0\}.
\end{aligned}
\end{equation}
The pseudo-label set used for student training becomes
\begin{equation}
    \hat{y}_i^{\mathrm{rec}}
    =
    \hat{y}_i
    \cup
    \mathcal{V}_i^{\mathrm{rec}}.
\end{equation}
The unlabeled detection loss is then computed with $\hat{y}_i^{\mathrm{rec}}$ instead of $\hat{y}_i$.

\textbf{Multi-View Hard Recovery.}
\label{app:completion_multiview_details}
Multi-view hard recovery adds a stability signal to the hard recovery scorer. The teacher is evaluated on $V$ content-preserving weak augmentations of the same image, following consistency-based teacher-student training~\cite{Sohn_2020_NeurIPS_FixMatch,Liu_2021_ICLR_UnbiasedTeacher}, and all predicted boxes are mapped back to the original image coordinates. 
In our implementation, each weak view $a_w$ applies only random horizontal flipping with probability $0.5$; if a flip is applied, predicted boxes are inverted back to the original image coordinates before cross-view matching. 
A rejected candidate gains support if candidates of the same predicted class appear at similar locations in other weak views.

Let $r_{ij}^{v}$ be candidate $j$ from weak view $v$ after coordinate inversion. Its cross-view support is
\begin{equation}
    u_{ij}^{v}
    =
    \frac{1}{V-1}
    \sum_{v'\neq v}
    \mathbf{1}
    \left[
    \max_{j':\hat{c}_{ij'}^{v'}=\hat{c}_{ij}^{v}}
    \operatorname{IoU}(\hat{\mathbf{b}}_{ij}^{v},\hat{\mathbf{b}}_{ij'}^{v'})
    \geq
    \eta_{\mathrm{mv}}
    \right].
\end{equation}
The maximum over an empty set is zero. The support value is appended to $\phi_{ij}$. The scorer is trained with the same recovery labels as in hard recovery. At inference time, selected candidates are inserted into the pseudo-label set after duplicate removal.

\textbf{Foreground Query Revival.}
\label{app:completion_revival_details}
Foreground query revival uses recovered candidates as weak foreground evidence rather than hard pseudo-labels. This variant keeps the pseudo-label set unchanged and adds a loss on the corresponding student predictions. It is designed to test whether rejected candidates can provide useful objectness supervision without requiring their class and box predictions to be trusted as full targets.

Let $\mathcal{V}_i$ be the selected recovery candidate set for image $i$. For each $r_{ij}\in\mathcal{V}_i$, $\mu(i,j)$ denotes the corresponding student prediction on the strong view $a_s(x_i)$.
In our implementation, $a_s$ is a photometric-only target view: brightness, contrast, and saturation are jittered in $[0.6,1.4]$, hue is jittered in $[-0.1,0.1]$, grayscale is applied with probability $0.2$, and Gaussian blur uses kernel size $5$ with $\sigma\in[0.1,2.0]$.
No geometric perturbation is used in this strong view, so the image coordinate system is unchanged.
The reported revival experiments match recovery candidates to student predictions by class-agnostic box overlap on this strong view, using IoU threshold $0.4$.

Let $\ell_{\mu(i,j)}$ be the class logits of the matched student prediction, and let $\operatorname{fg}(\cdot)$ aggregate foreground-class logits into a scalar foreground logit.
Although the code supports maximum, top-$2$ mean, and normalized log-sum-exp pooling, all reported experiments use normalized log-sum-exp,
\begin{equation}
    \operatorname{fg}(\ell)
    =
    T\log
    \left(
    \sum_{c=1}^{C}\exp(\ell_c/T)
    \right)
    -
    T\log C,
\end{equation}
with temperature $T=1$.

The revival loss is
\begin{equation}
    \mathcal{L}_{\mathrm{rev}}
    =
    \frac{1}{|\mathcal{V}_i|+\epsilon}
    \sum_{r_{ij}\in\mathcal{V}_i}
    \operatorname{BCE}_{\mathrm{logit}}
    \left(
    \operatorname{fg}(\ell_{\mu(i,j)}),
    a_{\mathrm{fg}}
    \right),
\end{equation}
where $\operatorname{BCE}_{\mathrm{logit}}$ denotes binary cross entropy applied to a logit.
The target value is $a_{\mathrm{fg}}=0.7$ in all reported revival runs.
The unlabeled objective becomes
\begin{equation}
    \mathcal{L}_{\mathrm{u}}^{\mathrm{rev}}
    =
    \mathcal{L}_{\mathrm{u}}
    +
    \lambda_{\mathrm{rev}}\mathcal{L}_{\mathrm{rev}}.
\end{equation}
The accepted pseudo-label set $\hat{y}_i$ remains unchanged.

\textbf{Multi-View Foreground Revival.}
\label{app:completion_multiview_revival_details}
Multi-view foreground revival combines the multi-view candidate scorer with the foreground revival loss. The candidate set $\mathcal{V}_i$ is selected using cross-view features, but the selected candidates are not inserted as full pseudo-labels. They only determine which student predictions receive the foreground revival loss. This variant separates the effect of multi-view candidate discovery from the risk of hard pseudo-label insertion.

\subsection{Optimization Control Details}
\label{app:optimization_details}

Optimization control uses sparse annotations after pseudo-labels have been constructed. The pseudo-label set remains unchanged. The three plugins instead regulate the relative scale of the unlabeled objective, the gradient interaction between sparse supervision and pseudo-label learning, or the teacher update that determines future pseudo-labels.

\textbf{Sparse Loss Balancing.}
\label{app:optimization_loss_balance_details}
Sparse loss balancing rescales the unlabeled branch according to its magnitude relative to sparse supervised learning.
Let $\bar{L}_{\mathrm{sup}}$ and $\bar{L}_{\mathrm{u}}$ be exponential moving averages of $\mathcal{L}_{\mathrm{sup}}$ and $\mathcal{L}_{\mathrm{u}}$. For a generic scalar loss $L_t$, the moving average is
\begin{equation}
    \bar{L}_t
    =
    \omega\bar{L}_{t-1}
    +
    (1-\omega)L_t.
\end{equation}
The same update is applied separately to the supervised and unlabeled losses. The unlabeled scale is defined as
\begin{equation}
    \lambda_t
    =
    \operatorname{clip}
    \left(
    \left(
    \frac{\kappa\bar{L}_{\mathrm{sup}}}{\bar{L}_{\mathrm{u}}+\epsilon}
    \right)^{\alpha},
    \lambda_{\min},
    \lambda_{\max}
    \right).
\end{equation}
$\kappa$ specifies the target ratio between unlabeled and supervised magnitudes, $\alpha$ controls the update strength, and the clipping range prevents unstable rescaling. The student objective becomes
\begin{equation}
    \mathcal{L}
    =
    \mathcal{L}_{\mathrm{sup}}
    +
    \lambda_t\mathcal{L}_{\mathrm{u}}.
\end{equation}
The default setting scales the full unlabeled branch. 

\textbf{Target-Anchored PCGrad.}
\label{app:optimization_pcgrad_details}
Target-anchored PCGrad treats the sparse supervised gradient as the trusted target direction. Let $g_s=\nabla_{\theta}\mathcal{L}_{\mathrm{sup}}$ be the gradient from sparse labeled target data and $g_p=\nabla_{\theta}\mathcal{L}_{\mathrm{p}}$ be the gradient from the hard pseudo-label detection loss inside the unlabeled branch. If $g_p^{\top}g_s<0$, the pseudo-label gradient contains a component that opposes sparse supervised learning. The plugin removes this component while leaving the supervised gradient unchanged,
\begin{equation}
    g_p'
    =
    g_p
    -
    \mathbf{1}[g_p^{\top}g_s<0]
    \frac{g_p^{\top}g_s}{\|g_s\|_2^2+\epsilon}
    g_s.
\end{equation}
The final update direction is
\begin{equation}
    g
    =
    g_s
    +
    g_p'
    +
    g_{\mathrm{aux}},
\end{equation}
where $g_{\mathrm{aux}}$ denotes gradients from auxiliary unlabeled objectives and is zero when no such objectives exist. If $\|g_s\|_2$ is numerically too small, the projection is skipped.

\textbf{Label-Guided Teacher Update.}
\label{app:optimization_teacher_details}
Label-guided teacher update uses sparse supervision to identify teacher parameters that should follow the student more quickly. Let $\bar{\theta}$ and $\theta$ denote teacher and student parameters. Let $\mathcal{P}$ index the parameter entries considered by the teacher update. For a sparse labeled batch $\mathcal{B}_l$, the target-supervised importance of parameter entry $p\in\mathcal{P}$ is
\begin{equation}
    I_l(p)
    =
    \left\|
    \nabla_{\bar{\theta}_p}
    \mathcal{L}_{\mathrm{sup}}(f_{\bar{\theta}},\mathcal{B}_l)
    \right\|.
\end{equation}
The importance values are normalized by their global mean magnitude. If the SFDA-OD baseline such as DDT~\cite{He_2025_ICCV_DDT} provides an unlabeled importance estimate $I_u(p)$, the two signals are merged as
\begin{equation}
    I(p)
    =
    \max
    \left(
    \operatorname{norm}(I_l(p)),
    \operatorname{norm}(I_u(p))
    \right).
\end{equation}
If no unlabeled importance is available, $I(p)=\operatorname{norm}(I_l(p))$.

Let $\zeta_{q}$ denote the cutoff value for the top $q_{\mathrm{imp}}$ fraction of merged importance scores. Parameter entries with importance above this cutoff use the faster EMA momentum $\mu_{\mathrm{fast}}$, while the remaining entries use the baseline momentum $\mu_{\mathrm{base}}$,
\begin{equation}
    \theta_{t,p}
    \leftarrow
    \mu_p \theta_{t,p}
    +
    (1-\mu_p)\theta_{s,p},
    \qquad
    \mu_p
    =
    \begin{cases}
    \mu_{\mathrm{fast}}, & I(p)\geq \zeta_q,\\
    \mu_{\mathrm{base}}, & \mathrm{otherwise}.
    \end{cases}
\end{equation}
The condition $\mu_{\mathrm{fast}}<\mu_{\mathrm{base}}$ makes selected entries follow the student more quickly. This plugin changes only the teacher update. The student loss and pseudo-label set remain unchanged.

\subsection{Hyperparameter Summary}
\label{app:plugin_hyperparameters}

Tables~\ref{tab:plugin_hyperparams_selection}, \ref{tab:plugin_hyperparams_completion}, and \ref{tab:plugin_hyperparams_optimization} list the hyperparameters introduced by the plugin families. Unless otherwise stated, the values are shared across datasets, detectors, SFDA-OD baselines, and random seeds. The tables report the experimental values, while the preceding subsections define the corresponding quantities.

\begin{table}[H]
\caption{Hyperparameters for pseudo-label selection plugins.}
\label{tab:plugin_hyperparams_selection}
\centering
\scriptsize
\setlength{\tabcolsep}{6pt}
\renewcommand{\arraystretch}{1.08}
\begin{tabular}{@{}lll@{}}
\toprule
Plugin & Paper variable & Value \\
\midrule
\multirow{2}{*}{Shared} & Verification IoU $\eta_{\mathrm{sel}}$ & $0.50$ \\
& Threshold range $[\tau_{\min},\tau_{\max}]$ & $[0.25,0.55]$ \\
\midrule
\multirow{3}{*}{Prior mapping} & Prior smoothing $\alpha_{\pi}$ & $1.0$ \\
& Ratio sensitivity $\beta$ & $0.75$ \\
& Offset bounds $(\delta_{\downarrow},\delta_{\uparrow})$ & $(0.10,0.10)$ \\
\midrule
\multirow{3}{*}{Precision calibration} & Precision target $p_{\min}$ & $0.75$ \\
& Support requirements $(n_{\min},m_{\min})$ & $(2,1)$ \\
& Offset bounds $(\delta_{\downarrow},\delta_{\uparrow})$ & $(0.10,0.15)$ \\
\midrule
\multirow{4}{*}{Reliability rescaling} & Precision target $p_{\min}$ & $0.75$ \\
& Support requirements $(n_{\min},m_{\min})$ & $(3,1)$ \\
& Score-weight range $[w_{\min},1]$ & $[0.50,1.00]$ \\
& Reliability exponent $\gamma$ & $1.0$ \\
\bottomrule
\end{tabular}
\end{table}

\begin{table}[H]
\caption{Hyperparameters for object completion plugins.}
\label{tab:plugin_hyperparams_completion}
\centering
\scriptsize
\setlength{\tabcolsep}{5.5pt}
\renewcommand{\arraystretch}{1.07}
\resizebox{\linewidth}{!}{
\begin{tabular}{@{}lll@{}}
\toprule
Plugin & Paper variable / design choice & Value \\
\midrule
\multirow{5}{*}{Shared}
& Coverage IoU $\eta_{\mathrm{cov}}$ & $0.50$ \\
& Recovery-label IoU $\eta_{\mathrm{rec}}$ & $0.50$ \\
& Duplicate-removal IoU $\eta_{\mathrm{dup}}$ & $0.70$ \\
& Recovery threshold $\rho$ & sparse-label precision estimate \\
& Strong student view $a_s$ & photometric only \\
\midrule
\multirow{2}{*}{Hard recovery}
& Recovery precision floor for $\rho$ & $0.55$ \\
& Weak teacher views $V$ & $1$ \\
\midrule
\multirow{4}{*}{Multi-view hard recovery}
& Recovery precision floor for $\rho$ & $0.50$ \\
& Weak teacher views $V$ & $2$ \\
& Additional weak view $a_w$ & hflip, $p=0.5$ \\
& Cross-view support IoU $\eta_{\mathrm{mv}}$ & $0.50$ \\
\midrule
\multirow{5}{*}{Foreground revival}
& Recovery setting & Hard Query Recovery \\
& Foreground target $a_{\mathrm{fg}}$ & $0.70$ \\
& Revival loss weight $\lambda_{\mathrm{rev}}$ & $0.02$ \\
& Foreground aggregation $\operatorname{fg}(\cdot)$ & mean-logsum-exp, $T=1.0$ \\
& Student matching & class-agnostic IoU $\geq0.40$ on $a_s(x)$ \\
\midrule
\multirow{2}{*}{Multi-view foreground revival}
& Recovery setting & same as Multi-View Hard Recovery \\
& Revival setting & same as Foreground Query Revival \\
\bottomrule
\end{tabular}}
\end{table}

\begin{table}[H]
\caption{Hyperparameters for optimization control plugins.}
\label{tab:plugin_hyperparams_optimization}
\centering
\scriptsize
\setlength{\tabcolsep}{6pt}
\renewcommand{\arraystretch}{1.08}
\begin{tabular}{@{}lll@{}}
\toprule
Plugin & Paper variable / design choice & Value \\
\midrule
\multirow{4}{*}{Loss balancing} & Loss EMA momentum $\omega$ & $0.95$ \\
& Target ratio $\kappa$ & $1.0$ \\
& Scale exponent $\alpha$ & $0.5$ \\
& Scale range $[\lambda_{\min},\lambda_{\max}]$ & $[0.5,1.5]$ \\
\midrule
\multirow{4}{*}{PCGrad} & Anchor gradient $g_s$ & Sparse supervised target loss \\
& Projected gradient $g_p$ & Hard pseudo-label loss only \\
& Auxiliary gradient $g_{\mathrm{aux}}$ & unchanged \\
& Numerical $\epsilon$ & $10^{-12}$ \\
\midrule
\multirow{5}{*}{Teacher update} & Importance merge $I(p)$ & $\max(\operatorname{norm}(I_l),\operatorname{norm}(I_u))$ \\
& Importance index set $\mathcal{P}$ & parameter entries \\
& Importance fraction $q_{\mathrm{imp}}$ & $0.10$ \\
& Faster EMA momentum $\mu_{\mathrm{fast}}$ & $0.997$ \\
& Baseline EMA momentum $\mu_{\mathrm{base}}$ & $0.999$ \\
\bottomrule
\end{tabular}
\end{table}

\section{Additional Results}
\label{app:add-exp}



\textbf{Additional Budget Results.}
Table~\ref{tab:budget_ablation_foggy_dino_ddt} reports the sparse-label budget ablation on Cityscapes $\rightarrow$ Foggy Cityscapes using DINO and DDT.
The table compares $1\%$, $5\%$, and $10\%$ random target-label budgets under the same RTSM and plugin protocol used in the main experiments.
It also reports the average plugin change over RTSM for each budget.
Overall, increasing the budget consistently strengthens the RTSM anchor, while sparse-label feedback does not produce a stable improvement over the anchor across budgets.

\begin{table}[htb]
    \centering
    \caption{
        Results on Cityscapes $\rightarrow$ Foggy Cityscapes with DINO and DDT \textbf{under 1\%, 5\%, and 10\% random target-label budget}.
        We report the mean AP50 with standard deviation.
        For plugin rows, $\Delta$ reports the AP50 difference from the matched RTSM anchor.
        The final row averages the AP50 means and $\Delta$ values of all ten plugin variants for each budget.
    }
    \label{tab:budget_ablation_foggy_dino_ddt}
    \vspace{-10pt}
    \scriptsize
    \setlength{\tabcolsep}{0.05cm}
    \renewcommand{\arraystretch}{1.08}
    \newcommand{\budgetapstd}[2]{#1{\scriptsize$_{\pm #2}$}}
    \newcommand{\budgetposdelta}[1]{\cellcolor[HTML]{DDF8CB}\textcolor[HTML]{1B6E1B}{+#1}}
    \newcommand{\budgetnegdelta}[1]{\cellcolor[HTML]{F8D7DA}\textcolor[HTML]{8A1C1C}{#1}}
    \begin{tabular}{@{}l*{3}{rc}@{}}
        \toprule
        \multirow{2}{*}{\begin{tabular}[c]{@{}l@{}}Protocol /\\ Plugin\end{tabular}} &
        \multicolumn{2}{c}{1\% Labels} &
        \multicolumn{2}{c}{5\% Labels} &
        \multicolumn{2}{c}{10\% Labels} \\
        \cmidrule(lr){2-3}\cmidrule(lr){4-5}\cmidrule(l){6-7}
        & AP50 & $\Delta$ & AP50 & $\Delta$ & AP50 & $\Delta$ \\
        \midrule
        \rowcolor[HTML]{F5F5F5}
        Source-Only & \multicolumn{6}{c}{32.608} \\
        \rowcolor[HTML]{F5F5F5}
        Full-Target Oracle & \multicolumn{6}{c}{62.507} \\
        \rowcolor[HTML]{F9F9F9}
        Pure SFDA & \multicolumn{6}{c}{\budgetapstd{53.669}{0.350}} \\
        \rowcolor[HTML]{F9F9F9}
        RTSM
        & \multicolumn{2}{c}{\budgetapstd{55.415}{1.754}}
        & \multicolumn{2}{c}{\budgetapstd{58.695}{0.644}}
        & \multicolumn{2}{c}{\budgetapstd{59.910}{0.672}} \\
        \midrule
        \rowcolor[HTML]{D3D3D3}
        \multicolumn{7}{@{}l}{\hspace{0.4em}\textbf{Pseudo-Label Selection}} \\
        Prior-Mapped Thresholding
        & \budgetapstd{55.918}{0.414} & \budgetposdelta{0.503}
        & \budgetapstd{59.005}{1.068} & \budgetposdelta{0.311}
        & \budgetapstd{59.471}{0.587} & \budgetnegdelta{-0.439} \\
        Precision-Calibrated Thresholding
        & \budgetapstd{56.674}{0.439} & \budgetposdelta{1.259}
        & \budgetapstd{58.631}{0.938} & \budgetnegdelta{-0.064}
        & \budgetapstd{59.745}{0.460} & \budgetnegdelta{-0.165} \\
        Reliability Score Rescaling
        & \budgetapstd{56.652}{0.648} & \budgetposdelta{1.237}
        & \budgetapstd{59.112}{0.242} & \budgetposdelta{0.418}
        & \budgetapstd{60.435}{0.336} & \budgetposdelta{0.525} \\
        \midrule
        \rowcolor[HTML]{D3D3D3}
        \multicolumn{7}{@{}l}{\hspace{0.4em}\textbf{Object Completion}} \\
        Hard Query Recovery
        & \budgetapstd{50.995}{4.348} & \budgetnegdelta{-4.420}
        & \budgetapstd{53.742}{0.764} & \budgetnegdelta{-4.953}
        & \budgetapstd{56.155}{1.289} & \budgetnegdelta{-3.755} \\
        Multi-View Hard Recovery
        & \budgetapstd{51.322}{0.866} & \budgetnegdelta{-4.093}
        & \budgetapstd{53.393}{1.851} & \budgetnegdelta{-5.302}
        & \budgetapstd{55.129}{1.222} & \budgetnegdelta{-4.781} \\
        Foreground Query Revival
        & \budgetapstd{56.593}{0.522} & \budgetposdelta{1.178}
        & \budgetapstd{59.012}{1.353} & \budgetposdelta{0.317}
        & \budgetapstd{59.527}{0.869} & \budgetnegdelta{-0.383} \\
        Multi-View Foreground Revival
        & \budgetapstd{56.742}{0.667} & \budgetposdelta{1.327}
        & \budgetapstd{58.954}{0.830} & \budgetposdelta{0.259}
        & \textbf{\budgetapstd{60.696}{0.272}} & \budgetposdelta{0.786} \\
        \midrule
        \rowcolor[HTML]{D3D3D3}
        \multicolumn{7}{@{}l}{\hspace{0.4em}\textbf{Optimization Control}} \\
        Sparse Loss Balancing
        & \budgetapstd{56.527}{0.420} & \budgetposdelta{1.112}
        & \budgetapstd{58.820}{1.017} & \budgetposdelta{0.125}
        & \budgetapstd{59.716}{0.518} & \budgetnegdelta{-0.194} \\
        Target-Anchored PCGrad
        & \budgetapstd{56.409}{0.379} & \budgetposdelta{0.994}
        & \textbf{\budgetapstd{59.147}{0.770}} & \budgetposdelta{0.452}
        & \budgetapstd{60.285}{0.629} & \budgetposdelta{0.375} \\
        Label-Guided Teacher Update
        & \textbf{\budgetapstd{57.290}{0.787}} & \budgetposdelta{1.875}
        & \budgetapstd{58.577}{1.387} & \budgetnegdelta{-0.118}
        & \budgetapstd{59.779}{0.273} & \budgetnegdelta{-0.132} \\
        \midrule
        \rowcolor[HTML]{EFEFEF}
        \textbf{Average of Plugins}
        & {55.512} & \budgetposdelta{0.097}
        & {57.839} & \budgetnegdelta{-0.856}
        & {59.094} & \budgetnegdelta{-0.816} \\
        \bottomrule
    \end{tabular}
\vspace{-15pt}
\end{table}

\end{document}